\documentclass[lettersize,journal]{IEEEtran}
\usepackage{amsmath,amssymb}
\usepackage{array}
\usepackage{footnote}
\usepackage{booktabs,makecell,tabularx}
\usepackage{multirow}
\usepackage{subcaption}
\usepackage{textcomp}
\usepackage{stfloats}
\usepackage{url}
\usepackage{verbatim}
\usepackage{graphicx}
\usepackage[mathscr]{euscript}
\usepackage{algorithm}%
\usepackage{algorithmicx}%
\usepackage{algpseudocode}%
\usepackage{listings}%
\usepackage[table]{xcolor}
\usepackage{hyperref}   

\def\BibTeX{{\rm B\kern-.05em{\sc i\kern-.025em b}\kern-.08em
    T\kern-.1667em\lower.7ex\hbox{E}\kern-.125emX}}
\usepackage{balance}
\definecolor{maroon}{cmyk}{0,0.87,0.68,0.32}
\definecolor{teal}{cmyk}{0,0.87,0.68,0.32}
\definecolor{ForestGreen}{RGB}{34,139,34}

\definecolor{Gray}{gray}{0.9}
\definecolor{LightCyan}{rgb}{0.85,1,1}

\begin{document}
\title{HMR-Net: Hierarchical Modular Routing for Cross-Domain Object Detection in Aerial Images}
\author{Pourya~Shamsolmoali,~\IEEEmembership{Senior Member,~IEEE,}
        Masoumeh~Zareapoor,~\IEEEmembership{Member,~IEEE,} Michael~Felsberg,~\IEEEmembership{Senior Member,~IEEE,} Nick~Pears, Huiyu~Zhou
        and~Yue~Lu,~\IEEEmembership{Senior Member,~IEEE}

\thanks{ P.~Shamsolmoali and N.~Pears are with the Department of Computer Science, University of York, UK. (\{pourya.shamsolmoali, nick.pears\}@york.ac.uk).}
\thanks{M.~Zareapoor is with the SEIEE, Shanghai Jiao Tong University, China. (mzarea222@gmail.com).}
\thanks{M.~Felsberg is with the Computer Vision Laboratory, Linkoping University, Sweden. (michael.felsberg@liu.se).}
\thanks{H.~Zhou is with the School of Computing and Mathematical Sciences, University of Leicester, UK. (hz143@leicester.ac.uk).}
\thanks{Y.~Lu is with the SCEE, East China Normal University, China. (ylu@cee.ecnu.edu.cn).}
}

\markboth{IEEE TRANSACTIONS ON GEOSCIENCE and REMOTE SENSING}%
{How to Use the IEEEtran \LaTeX \ Templates}

\maketitle

\begin{abstract}
Despite advances in object detection, aerial imagery remains a challenging domain, as models often fail to generalize across variations in spatial resolution, scene composition, and semantic label coverage. Differences in geographic context, sensor characteristics, and object distributions across datasets limit the capacity of conventional models to learn consistent and transferable representations. Shared methods trained on such data tend to impose a unified representation across fundamentally different domains, resulting in poor performance on region-specific content and less flexibility when dealing with novel object categories. To address this, we propose a novel modular learning framework that enables structured specialization in aerial detection. Our method introduces a hierarchical routing mechanism with two levels of modularity: a domain routing layer that uses latent geographic embeddings to assign inputs to domain-specialized expert modules, and a scene routing mechanism that allocates image subregions to scene-specific expert modules. This allows our method to specialize across datasets and within complex scenes. Additionally, the framework contains a conditional expert module that uses external semantic information (e.g., category names or textual descriptions) to enable detection of novel object categories during inference, without the need for retraining or fine-tuning.
By moving beyond monolithic representations, our method provides an adaptive framework for remote sensing object detection. Comprehensive evaluations on four datasets highlight improvements in multi-dataset generalization, region-level specialization, and open-category detection.
\end{abstract}


\section{Introduction}
\IEEEPARstart {O}{bject} detection in aerial or remote sensing imagery (RSI) refers to identifying objects of interest on the Earth surface \cite{shamsolmoali2023efficient, zhang2026mdadet}. This task is important for interpreting remote sensing data and supporting a range of earth observation applications, including environmental monitoring and disaster response systems \cite{deng2023towards, yu2026tsc}.
Recent advancements in deep learning, particularly with the rise of convolutional neural networks (CNNs), have significantly improved aerial object detection \cite{fu2026oaadet, deng2023towards, li2024lsknet, shamsolmoali2021rotation}. Despite these improvements, object detection in RSI continues to pose considerable challenges due to the substantial diversity in data sources and scene structures \cite{huang2025survey}.
Unlike natural image benchmarks \cite{lin2014microsoft, deng2009imagenet}, remote sensing datasets \cite{li2020object, xia2018dota, lam2018xview} are collected across different geographic regions and sensor platforms, each with its own characteristics in spatial resolution, semantic coverage, and object distribution. This heterogeneity introduces significant domain shifts, making it difficult for a unified model to capture generalizable representations across datasets \cite{liu2024source}.

\begin{figure}[t]
\centering
\includegraphics[width=0.99\columnwidth]{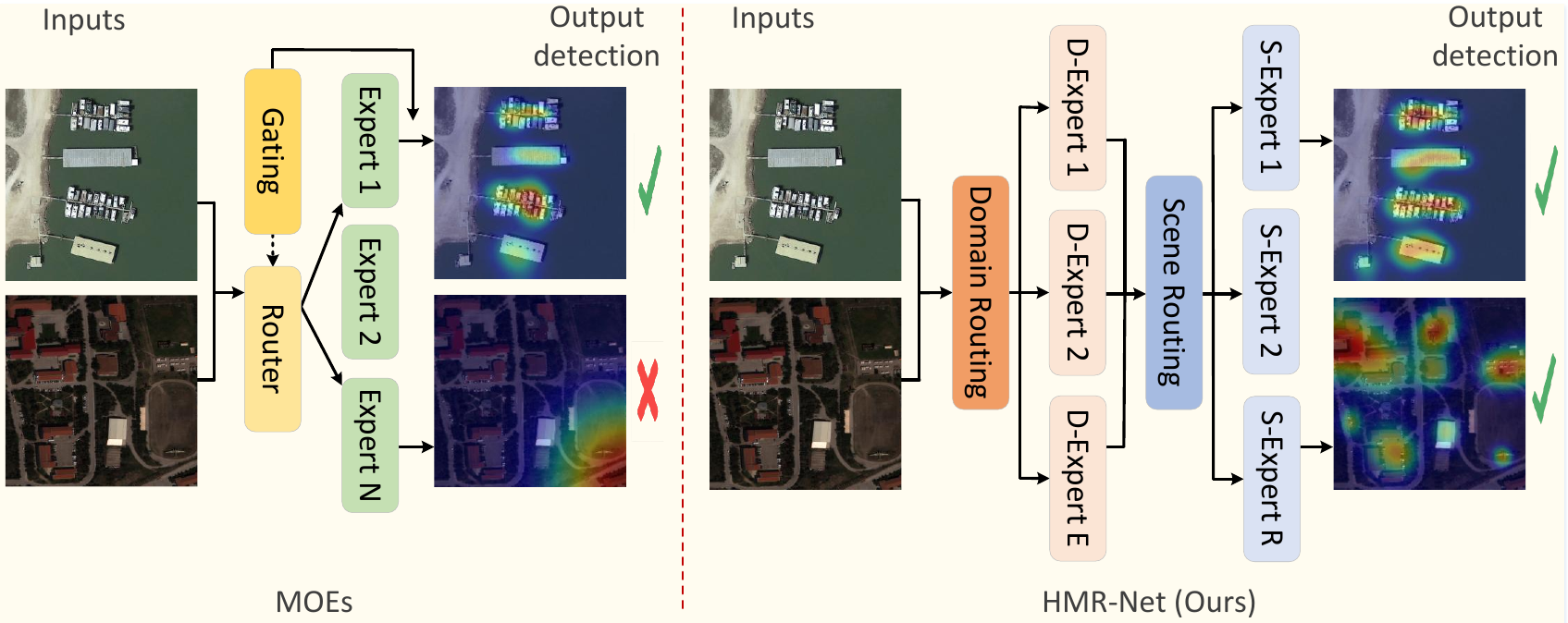} 
\vspace{-6pt}
\caption{\small{Comparison between standard Mixture of Experts (MOE) and our Hierarchical Modular Routing. Unlike conventional models with global-only expert assignment, our method introduces dataset-specific domain routing and region-level scene specialization.}}
\label{teaser}
\end{figure} 

A typical approach to address multi-domain object detection is training a shared model over concatenated datasets \cite{zhang2023nico++, bi2025good}. While this strategy provides a simple solution, it often results in biased learning, where dominant datasets or frequently occurring object types overshadow less represented domains. Such shared architectures lack the capacity to adapt to domain-specific variations or to focus selectively on geographically or structurally distinct regions within a scene.
To overcome these limitations, recent methods have explored multi-task learning \cite{liu2023hierarchical} and Mixture of Experts (MoE) \cite{jain2023damex, le2024mixture} frameworks. Multi-task approaches introduce dataset-specific components to encourage partial specialization, while MoE models introduce expert modules and gating functions that dynamically route inputs, allowing selective processing based on content. However, these models often operate at the token level with soft attention, resulting in fragmented or overlapping expert usage. Their gating mechanisms lack semantic structure, and expert usage is imbalanced. More importantly, as shown on Fig. \ref{teaser} (left) they ignore the spatial context inherent to remote sensing imagery, where object appearance is often determined by region-specific visual patterns.
Moreover, most models rely on fixed label spaces and cannot generalize to novel categories during inference \cite{zhang2023remote, zareapoor2025bimac}.

These challenges require a detection framework that can move beyond uniform representations and instead adapt structurally to the diversity of remote sensing data. Specifically, the framework should support specialization across heterogeneous datasets, adapt to spatial variability within individual scenes, and flexibly extend to novel categories at inference time without requiring manual reconfiguration or retraining.
To address these challenges, we propose Hierarchical Modular Routing Network (HMR-Net), that introduces structured, multi-level specialization into the object detection pipeline. Instead of relying on a single shared backbone, the model is composed of modular components that are selectively activated based on dataset identity and scene composition. As shown in Fig. \ref{teaser} (right), at the dataset level, we introduce a domain expert routing mechanism that learns latent geographic embeddings to represent each dataset and uses them to assign inputs to domain-specialized expert modules. This enables the model to capture coarse domain characteristics such as those present in urban, rural, or coastal imagery without relying on manual dataset structuring.

At the image level, we introduce a content-aware spatial parsing strategy that adaptively splits each image into region-specific subspaces, enabling targeted expert selection based on localized visual context. These region-level experts enable the model to capture fine-grained, spatially localized structures, supporting contextual adaptation within a single scene, such as roads in dense urban zones or vessels in coastal image. Together, domain routing and scene routing form the two levels of our hierarchical design, enabling inter-domain adaptability and intra-scene specialization within a unified framework. 

To enhance its flexibility, we extend this hierarchical structure by allowing expert behaviour to be conditionally guided by external semantic cues. Specifically, we introduce a mechanism for latent alignment between visual representations and text inputs, enabling the model to respond to object categories not present during training. This supports open-category inference through prompt adaptation, without requiring explicit supervision or retraining. In doing so, the method moves beyond fixed label spaces and acquires the flexibility needed for real-world remote sensing tasks.

Through this design, HMR-Net captures both domain-level variation across datasets and fine-grained spatial structure within scenes. Semantic guidance also supports the detection of previously unseen object categories without retraining. Extensive experiments on diverse benchmarks demonstrate consistent improvements in cross-domain generalization, localization accuracy, and open-category detection. These results highlight the effectiveness of hierarchical modular learning for object detection in complex remote sensing environments.
\noindent The main contributions are summarized as follows:
\begin{itemize}
\item We propose HMR-Net, a framework for RSI object detection that enables structured specialization at both the dataset and image-region levels through domain expert routing and scene-aware expert selection.
\item We further extend the modular framework with semantic conditioning for open-category detection.
\item Extensive experiments demonstrate consistent improvements in cross-domain generalization, region-level localization, and detection of previously unseen categories, highlighting the scalability and label efficiency of HMR-Net for remote sensing applications.
\end{itemize} 
The rest of this manuscript is structured as follows: Section \ref{sec2} discusses related studies. In Section \ref{sec3}, we introduce the details of HMR-Net. Section \ref{sec4} presents the experimental results. Finally, Section \ref{sec5} concluding the remarks.

\section{Related Work}\label{sec2}

\subsection{Object Detection in Remote Sensing}
Object detection in remote sensing has been addressed through a range of deep learning approaches adapted from general-purpose detection models \cite{shamsolmoali2021multipatch, li2024lsknet, yang2022h2rbox}, including two-stage detectors like Faster R-CNN \cite{xie2021oriented} and one-stage variants such as YOLO \cite{lin2023yolo, wei2025sed} and RetinaNet \cite{he2024robustness, miao2022improved}. Moreover, attention \cite{sun2020sraf, shamsolmoali2023efficient}, transformer \cite{zhu2023transformer, zhao2024orientedformer} and feature enhancement-based methods \cite{liu2024tiny, lyu2022rtmdet} have also been extended to this domain. These methods have shown strong performance on individual benchmarks such as DOTA, xView, and DIOR \cite{xia2018dota, lam2018xview, li2020object}, which vary widely in resolution, object density, and scene type.
Despite these advances, most models are developed and evaluated on a per-dataset basis and are rarely designed to operate across multiple sources \cite{li2020object, azimi2018towards, li2024lsknet}. Cross-dataset generalization remains limited, as models tend to overfit to dataset-specific characteristics and struggle with variations in scene structure and label semantics. Several works have attempted to improve transferability by applying domain adaptation techniques \cite{luo2024self, ma2024hierarchical} or training unified models over merged datasets \cite{zhang2023nico++, liu2022open}. However, these approaches often rely on shared backbones and global representations, which are insufficient to capture the spatial and semantic diversity inherent to remote sensing.

This has motivated the need for detection frameworks that can explicitly model domain-specific variations, scene spatial structures, and evolving object semantics challenges that remain insufficiently addressed by current approaches.

\subsection{Multi-Dataset Object Detection}
Object detection has long been a foundational task in computer vision, with continued efforts aimed at developing models that generalize across a wide range of object categories and visual environments. One of the key challenges in this area is the limited availability of large-scale, well-annotated datasets, as collecting and labeling high-quality data is both expensive and time-consuming \cite{oksuz2023mocae}. To address this limitation, recent research has explored the use of multiple datasets in a unified training framework, aiming to use the visual diversity across sources to improve robustness and generalization in detection tasks \cite{meng2023detection, shi2024plain}.
Training object detectors using multiple datasets is a widely adopted strategy for improving generalization, especially where individual datasets lack sufficient diversity or scale. By combining datasets, models are exposed to a wider range of object categories, imaging conditions, and domain variations. However, differences in dataset size, annotation coverage, label semantics, and scene structure introduce significant challenges.
Early studies focused on jointly training large datasets such as ImageNet and OpenImages \cite{yang2019detecting} to increase visual diversity and category coverage. However, differences in annotations and domain characteristics limited their effectiveness. To address these issues, methods such as squeeze-and-excitation layers were introduced to adapt feature representations to different datasets \cite{wang2019towards}. However, this approach still relied on shared architectures and did not address semantic inconsistencies across datasets. More recent methods tackle these challenges through partial-label learning, pseudo-labeling, and language-guided label-space alignment \cite{meng2023detection, shi2024plain}.

Other approaches use unified architectures with dataset-specific output and projection layers to reconcile heterogeneous label definitions during training \cite{zhou2022simple}. Mixture-of-experts models further improve specialization by assigning datasets to dedicated experts, enhancing robustness to domain imbalance and distribution shift \cite{jain2023damex, jin2024moe++, oksuz2023mocae}. However, these methods typically rely on fixed expert assignments, limiting flexibility when dataset boundaries are ambiguous or distributions overlap. In contrast, HMR-Net enables adaptive expert specialization at both the dataset and region levels through learned routing based on domain and spatial context.

\begin{figure}[t]
\centering
\includegraphics[width=0.99\columnwidth]{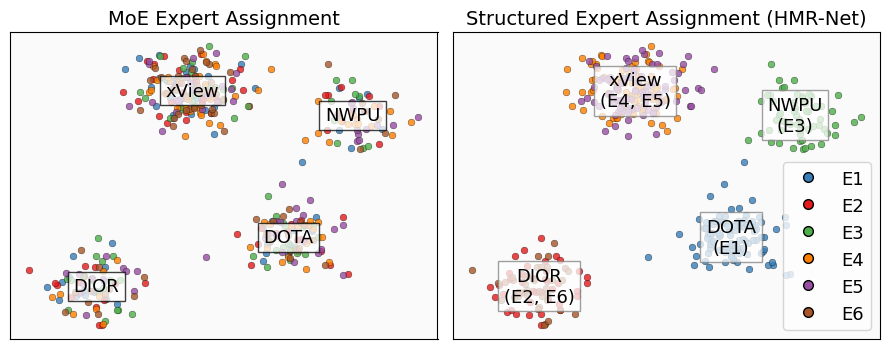} 
\vspace{-6pt}
\caption{\small{Visualization of domain expert assignments across datasets in latent space. Each point represents an image assignment projected into 2D using MDS. Left: MoE shows diffuse expert usage across datasets. Right: HMR-Net forms well-separated clusters, with experts specializing in different datasets and domains. 
}}
\label{exp-assign}
\end{figure} 
\subsection{Modular and Expert-Based Learning}
Recent work in modular and expert-based learning \cite{mittal2022modular, al2024mixture} has explored architectures that improve adaptability and specialization across tasks and domains. Among these, Mixture-of-Experts (MoE) models route inputs to specialized modules, enabling dynamic allocation of model capacity. Existing MoE methods employ routing strategies such as Top-k token routing \cite{lepikhin2020gshard, shazeer2017outrageously}, expert-driven token selection \cite{zhou2022simple, zhou2022mixture}, and expert refinement mechanisms \cite{clark2022unified, liu2022sparsity}. A key challenge is imbalanced expert utilization, where a small subset of experts receives most inputs. To address this issue, balancing losses and calibrated expert learning have been proposed to encourage more uniform and reliable expert behavior \cite{fedus2022switch, huang2024harder, aimar2023balanced, sanchez2024flexible}.
Despite their success, existing approaches exhibit important limitations, particularly in remote sensing. Most rely on probabilistic gating, leading to diffuse expert assignments in which multiple experts contribute without developing clear specialization. As illustrated in Fig.~\ref{exp-assign} (left), expert usage remains scattered across domains, reducing modularity and interpretability. Furthermore, most MoE models operate at the token or patch level without incorporating higher-level context such as domain identity, scene composition, or geographic information. As a result, they struggle to capture both domain-level variation and fine-grained spatial structure. These limitations motivate a more structured and context-aware framework for cross-domain remote sensing object detection.

\section{Methodology}\label{sec3}

\subsection{Problem Overview}

Consider a collection of $N$ aerial image datasets $\mathcal{D}={D_1,D_2,\dots,D_N}$, representing different geographic regions or imaging conditions. Each dataset $D_i$ contains images $I_j^i \in \mathbb{R}^{H \times W \times 3}$ with corresponding bounding-box and category annotations. The goal of multi-domain object detection is to learn a unified detector $f_{\theta}$, parameterized by $\theta$, that accurately predicts object categories and bounding boxes across datasets despite variations in scene composition, spatial resolution, object scale, and data distribution.
Formally, given an image $I_j^i$ from the dataset $D_i$, the detector produces predictions: 
\begin{equation}
    f_{\theta}(I_j^i) = \{(\mathbf{b}_k, c_k)\}_{k=1}^{K_j^i},
    \label{eq1}
\end{equation}
where each $\mathbf{b}_k \in \mathbb{R}^4$ is the coordinates of a detected bounding box and $c_k \in \{1,\ldots,C\}$ denotes the predicted object class.
The standard approach for multi-dataset training is to minimize the detection loss computed across all datasets, which can be written as
\begin{equation}
    \mathcal{L}_{\text{det}}(\theta) = \sum_{i=1}^{N}\sum_{j=1}^{|D_i|}\ell(f_{\theta}(I_j^i), Y_j^i),
    \label{eq2}
\end{equation}
where $Y_j^i$ denotes the ground-truth labels and $\ell(\cdot)$ combines classification and bounding box regression losses. While this objective is effective for standard object detection, optimizing it across heterogeneous domains often encourages the model to learn domain-agnostic representations that overlook important domain-specific characteristics. As a result, the shared parameters tend to encode overly generic features, limiting the model's ability to adapt to visually distinct or under-represented domains and ultimately degrading detection performance.
Another limitation of conventional detection architectures is their reliance on a single set of shared parameters to process data from diverse sources. Such a design restricts the model's capacity to capture domain-specific patterns and fine-grained spatial variations that are critical for robust detection across domains. In addition, most existing detection frameworks are designed for a fixed set of object categories, preventing them from recognizing unseen classes during inference.
\begin{figure}[t]
\centering
\includegraphics[width=0.99\columnwidth]{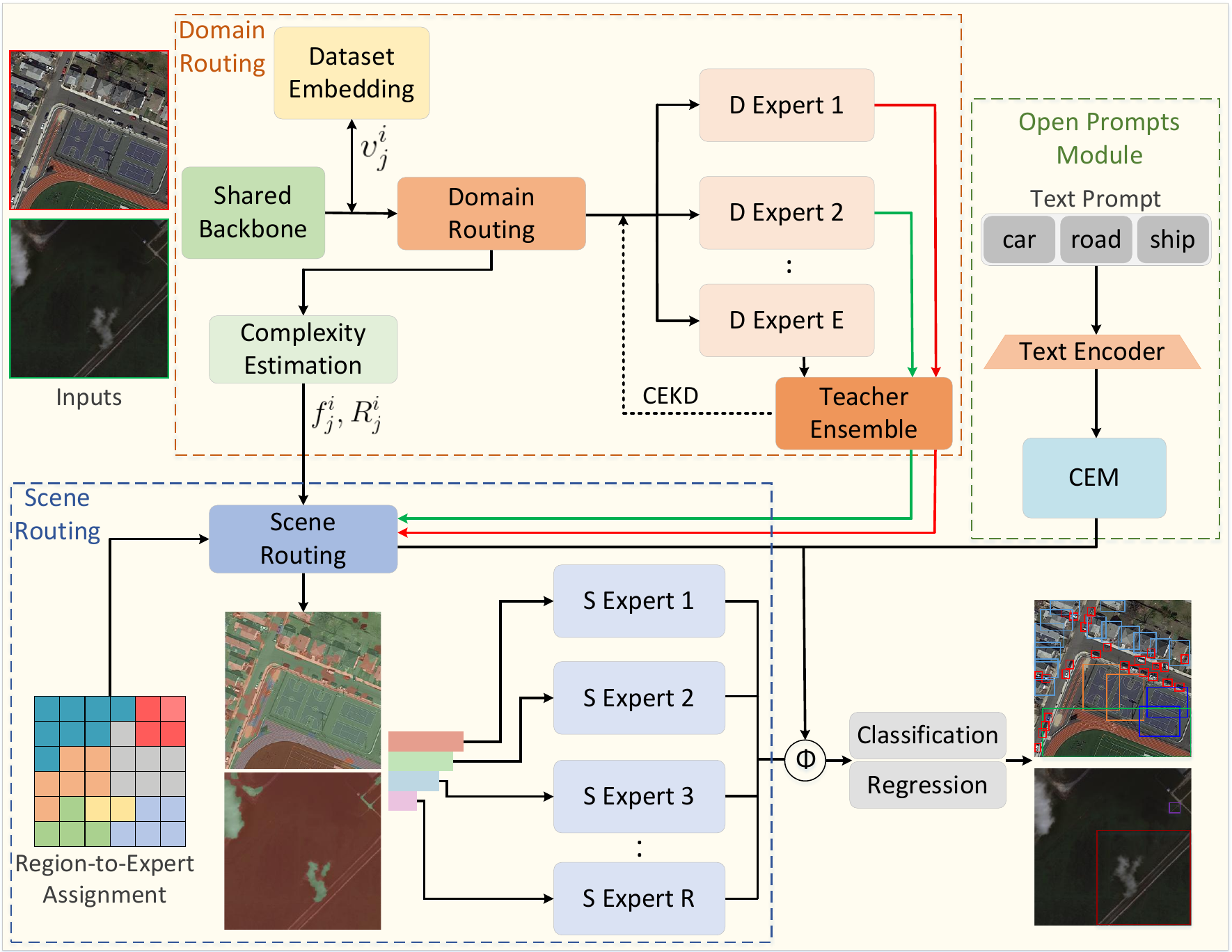} 
\vspace{-6pt}
\caption{\small{Overview of the proposed framework. Each image is routed to a domain expert, while a teacher ensemble of all experts provides cross-expert distillation. The routed output is partitioned into spatial regions and assigned to scene experts through complexity-aware routing for fine-grained specialization. Aggregated features are passed to detection heads, with an optional module for open-category detection.}}
\label{arch}
\end{figure} 

To address these challenges, we propose a Hierarchical Modular Routing Network (HMR-Net) that enables domain- and region-aware specialization through structured expert assignment, as illustrated in Fig.~\ref{exp-assign} (right). Unlike conventional MoEs, HMR-Net produces well-separated latent clusters, with experts specializing in different datasets and domains. The overall architecture is shown in Fig.~\ref{arch}.
\subsection{Domain Expert Routing}

To address domain variability in multi-domain object detection, we introduce a domain routing mechanism that assigns images to specialized experts based on dataset-level information. Unlike conventional MoEs \cite{al2024mixture}, which rely only on token-level visual features, our approach incorporates dataset-specific attributes to enable interpretable and domain-aware specialization.

Formally, given a collection of datasets $\mathcal{D}$, we associate each dataset $D_i$ with a latent embedding vector $\mathbf{z}_i \in \mathbb{R}^{d}$. Learned jointly with the model, these embeddings capture dataset-level priors, including semantic composition, geographic patterns, and overall visual characteristics. To improve routing stability, $\mathbf{z}_i$ is initialized using average pooled backbone features from each dataset and subsequently refined during training.
For each input image $I_j^i \in D_i$, we first extract an intermediate visual representation using a shared backbone $\phi(\cdot)$,
\begin{equation}
    \mathbf{v}_j^i = \phi(I_j^i) \in \mathbb{R}^{h \times w \times c}.
    \label{eq3}
\end{equation}
\noindent To determine dataset-level expert assignments, we summarize the feature map into a global descriptor by averaging over the spatial dimensions of the image \(\psi(\cdot)\), resulting in: $\mathbf{u}_j^i = \psi(\mathbf{v}_j^i) \in \mathbb{R}^{c}$. This descriptor is then combined with the corresponding dataset embedding $\mathbf{z}_i \in \mathbb{R}^{d}$ and passed to a domain routing function $G(\cdot)$ that outputs a probability distribution over the set of domain experts 
\begin{equation}
    \mathbf{g}_j^i = G([\mathbf{u}_j^i, \mathbf{z}_i]) \in \mathbb{R}^{E_{G}},
    \label{eq4}
\end{equation}
\noindent where $[\cdot,\cdot]$ denotes vector concatenation and $E_G$ is the number of domain experts. Unlike conventional gating strategies, the routing function conditions on the dataset embedding $\mathbf{z}_i$ to perform domain-aware expert assignment. The routing function is implemented as
\begin{equation}
\begin{aligned}
\tilde{x}_{ij} &= \mathrm{LN}\big([u^i_j, z_i]\big),\\
h_{ij} &= \phi\!\big(\tilde{x}_{ij} W_1 + b_1\big),\\
\mathbf{g}^i_j &= \sigma\!\left(\big(\tilde{x}_{ij} W_{\text{skip}} + h_{ij} W_2 + b_2\big)/\tau\right),
\end{aligned}
\label{5}
\end{equation}

\noindent in which $\phi$ denotes ReLU, $\sigma$ is the softmax over the $E_G$ experts, $\tau>0$ is a temperature parameter, and $W_{\text{skip}}, W_1, W_2, b_1,$ and $b_2$ are learnable parameters.
The concatenated representation $[\mathbf{u}_j^i, \mathbf{z}_i]$ enables the router to jointly consider image-level appearance and dataset context, producing a domain expert assignment vector $\mathbf{g}_j^i \in \mathbb{R}^{E_G}$.

To enforce discrete and interpretable specialization, we assign each input to the expert with the highest routing probability, $e_j^i = \arg\max_{e} \mathbf{g}_{j,e}^i$. Once the expert assignment $e_j^i$ is determined, the corresponding domain expert module $f_{e_j^i}^{\text{domain}}(\cdot)$ further processes the backbone representation $\mathbf{v}_j^i$, resulting in expert-specialized feature maps for detection
\begin{equation}
    \mathbf{f}_j^i = f_{e_j^i}^{\text{domain}}(\mathbf{v}_j^i).
    \label{eq6}
\end{equation}

\noindent To regularize specialization and minimize redundancy across domain experts, we use a cross-expert knowledge distillation (CEKD) strategy. For each input, we compute a softened ensemble output across all domain experts
\begin{equation}
    \mathbf{p}^{\text{teacher}} = \sigma\left( \frac{1}{E_{G}} \sum_{e=1}^{E_{G}} f_e^{\text{domain}}(\mathbf{v}_j^i) / T \right),
    \label{eq:teacher}
\end{equation}
\noindent where $T > 1$ is a temperature parameter and compute the student distribution from the routed expert
\begin{equation}
    \mathbf{p}^{\text{student}} = \sigma\left( f_{e_j^i}^{\text{domain}}(\mathbf{v}_j^i) / T \right).
    \label{eq:student}
\end{equation}

\noindent The distillation loss is then defined as the KL divergence between the teacher and student distributions
\begin{equation}
    \mathcal{L}_{\text{distill}} = \text{KL} \left( \mathbf{p}^{\text{teacher}} \, \| \, \mathbf{p}^{\text{student}} \right).
    \label{eq:distill}
\end{equation}
\noindent This regularization encourages each domain expert to remain consistent with the expert pool while preserving its specialization. In addition, $\mathcal{L}_{\text{distill}}$ aligns the outputs of domain experts into a shared representation space, enabling more stable scene expert processing and improved generalization across ambiguous domains.
\subsection{Scene Expert Routing}
Aerial images often contain multiple spatial regions with distinct semantic and structural characteristics, such as urban areas, vegetation, and water bodies. Domain experts alone cannot fully capture such intra-image variability, as they operate at the image level and do not explicitly model fine-grained spatial structure within individual scenes.
To address this limitation, we introduce a scene expert routing mechanism as the second stage of HMR-Net. This module partitions the domain expert feature maps into spatial {\emph{routing units}} (RUs) and routes each RU to an appropriate scene expert. The number of regions is determined according to scene complexity, enabling a balance between regional specialization and feature representation.

\noindent \textbf{Terminology.} We refer to the spatial partitions as RUs to distinguish them from \emph{object proposals (RoIs)} produced by the detector. Unless otherwise stated, "region" refers to an RU.

Given the domain expert feature representation $\mathbf{f}_j^i \in \mathbb{R}^{h \times w \times c'}$ from \eqref{eq6}, we first estimate scene complexity to determine the number of regions \(R_j^i\). Specifically, the entropy of \(\mathbf{f}_j^i\) is used as a lightweight measure of scene complexity:
\begin{equation}
\text{entropy}(\mathbf{f}_j^i) = -\sum_{c} p_c \log p_c, \quad p_c = \frac{\sum_{x,y} \mathbf{f}_j^i(x,y,c)}{\sum_{x,y,c'} \mathbf{f}_j^i(x,y,c')},
    \label{eq6-1}
\end{equation}
here $p_c$ is defined by normalizing the non-negative activations across channels \(c\), ensuring $p_c \in [0,1]$ and $\sum_c p_c=1$. Then we estimate the number of regions as
\begin{equation}
R_j^i = \left\lfloor \, 2 + \sigma_s\!\big(\theta_R \cdot \mathrm{entropy}(\mathbf{f}_j^i)\big)\,(E_{S} - 2) \, \right\rceil ,
    \label{eq6-2}
\end{equation}
where $\theta_R$ is a learnable scalar, $E_S$ is the maximum number of scene experts (set to 8 in our experiments), $\sigma_s$ denotes the sigmoid function, and $\lfloor \cdot \rceil$ denotes rounding to the nearest integer. This formulation ensures that $R_j^i$ is an integer, assigns at least two regions to each image, and allocates additional regions as scene complexity increases. The spatial partitioning map is then computed as
\begin{equation}
    \mathbf{P}_j^i = \Omega(\mathbf{f}_j^i;\theta_{\Omega}) \in \mathbb{R}^{h \times w \times E_{S}},
    \label{eq7}
\end{equation}
where $\Omega(\cdot)$ is a convolutional layer that produces an RU score tensor. 
We apply a softmax along the channel dimension so that, for each pixel $(x,y)$, the routing scores across all experts ($E_S$) sum to one. 
Then we select the top-$R_j^i$ channels ranked from their global score $\sum_{x,y}\mathbf P_j^i(x,y,r)$, leading to the candidate set of active experts. 
Therefore, the RU assignment for a spatial position \((x, y)\) is
\begin{equation}
    r_{x,y} = \arg\max_{r\in \mathcal{R}_j^i} \mathbf{P}_j^i(x,y,r),
    \label{eq8}
\end{equation}
where $\mathcal{R}_j^i$ denotes the index set of the top-$R_j^i$ experts. 
Subsequently, for each $r \in \{1,...,{R_j^i}\}$, the corresponding features from the domain expert feature map are aggregated into the RU-specific representations, which can be written as
\begin{equation}
    \mathbf{f}_{j,r}^i = \Gamma(\mathbf{f}_j^i, \mathbf{M}_{j,r}^i), \quad\quad \mathbf{M}_{j,r}^i(x,y)=
\begin{cases}
1 & \text{if } r_{x,y}=r\\[2pt]
0 & \text{otherwise},
\end{cases}
    \label{eq9}
\end{equation}
in which $\Gamma(\cdot)$ denotes a masked pooling operation that aggregates features within each RU. Using the binary mask $\mathbf{M}*{j,r}^i$, it pools the assigned spatial locations to produce a localized representation $\mathbf{f}*{j,r}^i$.
Each representation $\mathbf{f}_{j,r}^i$ is then routed to a scene expert module $f_r^{\text{scene}}(\cdot)$, which processes the localized features and produces region-specific outputs
\begin{equation}
    \mathbf{o}_{j,r}^i = f_{r}^{\text{scene}}(\mathbf{f}_{j,r}^i).
    \label{eq10}
\end{equation}
\noindent $f_{r}^{\text{scene}}$ is the $r$-th scene expert among the $E_S$ available experts. Only the top-$R_j^i$ experts selected by $\Omega$ are activated for each image. The outputs of the activated experts are fused into a unified feature representation, $
\mathbf{o}*{j}^{i} = \Phi\big({\mathbf{o}*{j,r}^{i}}_{r=1}^{R_j^i}\big)$, which is passed to the detection heads. Here, $\Phi(\cdot)$ denotes a spatially aligned fusion operation that combines region-specific outputs while preserving spatial consistency. To stabilize scene routing, we introduce the following regularization loss
\begin{equation}
    \mathcal{L}_{\text{coh}} = \!\!\!\!\!\!\!\sum_{(x,y) \sim (x',y')}\!\!\!\!\!\!\! \|\mathbf{P}_j^i(x, y) - \mathbf{P}_j^i(x', y')\|_2^2,
    \label{eq10-1}
\end{equation}
\noindent where \((x,y) \sim (x',y')\) denotes neighbouring pixels.
Region masks are obtained by applying $\arg\max$ to the top-$R_j^i$ maps in (\ref{eq8}), producing compact partitions, while (\ref{eq10-1}) enforces spatial continuity to ensure contiguous regions. In this process the computational cost is minimal, as entropy estimation and region prediction are lightweight operations. By combining scene routing with domain expert assignment, HMR-Net captures both regional and domain-level variations.
\begin{figure}
\centering
\includegraphics[width=0.99\columnwidth]{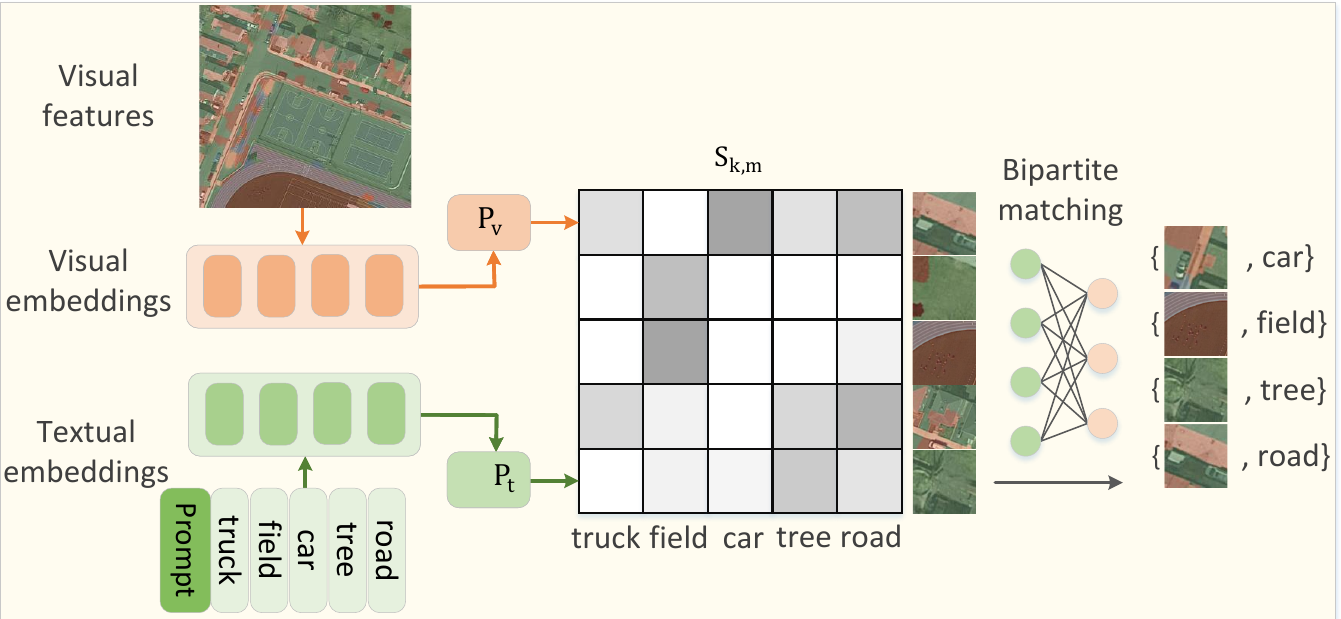} 
\vspace{-5pt}
\caption{\small{Conditional Expert Module: RU-level visual embeddings are aligned to textual prompts via bipartite matching, enabling prompt-driven open-category detection.}}
\label{arch2}
\end{figure} 
\subsection{Conditional Expert Module}
Open-vocabulary object detection aims to recognize object categories that are absent from the training set. Early approaches \cite{bansal2018zero} replaced the final classification layer with language embeddings, such as GloVe \cite{pennington2014glove}. More recent methods, including ViLD \cite{gu2021open}, Detic \cite{zhou2022detecting}, DetPro \cite{du2022learning}, AlignDet \cite{zareapoor2024learning}, and CastDet \cite{li2026exploiting}, use CLIP \cite{radford2021learning} to improve visual-language alignment. For example, ViLD distils region-level knowledge from CLIP image representations, while CastDet uses a CLIP-guided teacher-student framework.

Building on the domain-aware and region-specific representations learned through domain and scene routing, we introduce a Conditional Expert Module (CEM) for open-category detection. The CEM establishes a one-to-one correspondence between region-level visual embeddings and textual prompts derived from class names or descriptive phrases that capture contextual relationships in aerial scenes (see Fig. \ref{arch2}).

Let $\{\mathbf{v}_{j,r}^{i}\}_{r=1}^{R_j^i}$ be the set of RU-level visual embeddings for image $I_j^i$, and $\{\mathbf{t}_m\}_{m=1}^M$ be textual prompt embeddings generated by a frozen CLIP encoder \cite{radford2021learning}.
These are projected into a shared latent space using learnable projection functions
\begin{equation}
\tilde{\mathbf{v}}_{j,r}^{i} = \mathcal{P}_v(\mathbf{v}_{j,r}^{i}), \quad \tilde{\mathbf{t}}_m = \mathcal{P}_t(\mathbf{t}_m).
\label{C-1}
\end{equation}
where $\mathcal{P}_v(.)$ and $\mathcal{P}_t(.)$ are the projection function for the visual and textual domains, respectively. Between region $r$ and prompt $m$, we define the pairwise alignment score using similarity matrix $\mathbf{S}$ as
\begin{equation}
\mathbf{S}_{r,m} = \tilde{\mathbf{v}}_{j,r}^{i} \cdot \tilde{\mathbf{t}}_m.
\label{C-2}
\end{equation}
Our alignment loss encourages high similarity to correctly match pairs and is defined over a binary assignment matrix $\pi \in \{0,1\}^{{R_j^i} \times M}$, which can be defined by
\begin{equation}
\mathcal{L}_{\text{align}} = - \sum_{r,m} \mathbf{S}_{r,m} \cdot \pi_{r,m}.
\label{C-3}
\end{equation}
This loss is subject to a one-to-one matching. Each region must be matched to one prompt. Therefore, the optimal assignment $\hat{\pi}$ must maximize the total alignment score
\begin{equation}
\hat{\pi} = \arg\max_{\pi \in \Pi} \sum_{r,m} \mathbf{S}_{r,m} \cdot \pi_{r,m},
\label{C-4}
\end{equation}
where $\Pi$ denotes the set of all feasible one-to-one assignments. This bipartite matching problem is efficiently solved using the Hungarian algorithm \cite{kuhn1955hungarian}. The CEM operates on RU-level descriptors rather than RoI features. RoIs are generated after RU fusion by the detection head.

The CEM introduces minimal computational overhead, using a frozen CLIP text encoder and performing bipartite matching over a small number of visual-text pairs. At most 16 alignments are processed per image, resulting in less than $2\%$ additional inference time (see Fig.~\ref{N-image}). The aligned features are then passed to lightweight classification and bounding-box regression heads. The classification head supports both closed-set and prompt-guided detection.


\begin{algorithm}[t]
\caption{HMR-Net Training}
\label{alg:hmr}
\begin{algorithmic}[1]
\State \textbf{Init} dataset embeddings $\{\mathbf z_i\}$ by pooled backbone features; then train jointly.
\State \textbf{Note:} The number of routing units (RUs) per image satisfies $R_j^i \le E_S$.
\For{minibatch $\{(I_j^i,\text{labels})\}$}
  \State $\mathbf v_j^i \!\gets\! \phi(I_j^i)$;\quad $\mathbf u_j^i \!\gets\! \psi(\mathbf v_j^i)$
  \State $\mathbf g_j^i \!\gets\! \sigma(\mathbf W_g[\mathbf u_j^i,\mathbf z_i] + \mathbf{b}_g)$;\quad $e_j^i \!\gets\! \arg\max_{e\le E_G}\, \mathbf g_{j,e}^i$
  \State $\mathbf f_j^i \!\gets\! f_{e_j^i}^{\text{domain}}(\mathbf v_j^i)$ \hfill \eqref{eq6}
  \State Compute $\mathrm{entropy}(\mathbf f_j^i)$ and $R_j^i$ \hfill \eqref{eq6-1}, \eqref{eq6-2}
  \State $\mathbf P_j^i \!\gets\! \Omega(\mathbf f_j^i;\theta_\Omega)$; keep top-$R_j^i$ channels among $E_S$ \hfill \eqref{eq7}
  \State \textit{RU assignment:} $r_{x,y}\!\gets\!\arg\max_{r\le R_j^i}\mathbf P_j^i(x,y,r)$;\; build masks $\mathbf M_{j,r}^i$ \hfill \eqref{eq8}
  \For{$r=1..R_j^i$}
     \State $\mathbf f_{j,r}^i\!\gets\!\Gamma(\mathbf f_j^i,\mathbf M_{j,r}^i)$;\quad $\mathbf o_{j,r}^i\!\gets\! f_r^{\text{scene}}(\mathbf f_{j,r}^i)$ \hfill \eqref{eq9}, \eqref{eq10}
  \EndFor
  \State $\mathbf o_j^i \!\gets\! \Phi\!\big(\{\mathbf o_{j,r}^i\}_{r=1}^{R_j^i}\big)$;\; compute detection loss $\mathcal L_{\text{det}}$ on $\mathbf o_j^i$
  \State $\mathbf p^{\text{teacher}}$, $\mathbf p^{\text{student}}$ \hfill \eqref{eq:teacher}, \eqref{eq:student}
  \State $\mathcal L_{\text{coh}} \!\gets\! \sum_{(x,y)\sim(x',y')} \|\mathbf P_j^i(x,y)-\mathbf P_j^i(x',y')\|_2^2$ \hfill \eqref{eq10-1}
  \State Compute routing regularization $\mathcal L_{\text{route}}$ \hfill \eqref{eq16}
  \If{CEM enabled}
     \State Form RU embeddings $\{\mathbf v_{j,r}^{i}\}_{r=1}^{R_j^i}$ and prompts $\{\mathbf t_m\}_{m=1}^M$; project with $\mathcal P_v,\mathcal P_t$; build $\mathbf S$; $\hat\pi\!\gets\!\mathrm{Hungarian}$;
     \Statex \hspace{1.9em} $\mathcal L_{\text{align}}\!\gets\!-\sum_{r,m}\mathbf S_{r,m}\hat\pi_{r,m}$ \hfill \eqref{C-3}, \eqref{C-4}
  \Else
     \State $\mathcal L_{\text{align}}\!\gets\!0$
  \EndIf
  \State $\mathcal L \!\gets\! \mathcal L_{\text{det}} + \lambda_1 \mathcal L_{\text{route}} + \lambda_2 \mathcal L_{\text{distill}} + \lambda_3 \mathcal L_{\text{align}} + \lambda_4 \mathcal L_{\text{coh}}$; update parameters \hfill \eqref{eq:total_loss}
\EndFor
\end{algorithmic}
\end{algorithm}
\subsection{Training}
HMR-Net is trained using a multi-component objective that combines the detection loss with auxiliary regularization for expert routing, cross-expert consistency and visual-semantic alignment. To ensure stable optimization and effective expert specialization, we adopt a staged training strategy.

\noindent{\bf{Detection Loss:}}  
We use an object detection loss that combines classification and bounding box regression. For each dataset, the detection loss is defined as
\begin{equation}
    \mathcal{L}_{\text{det}} = \sum_{i=1}^{N} \sum_{j=1}^{M_i} \ell(\hat{Y}_j^i, Y_j^i),
    \label{eq15}
\end{equation}
where $\hat{Y}_j^i$ denotes model predictions for image $I_j^i$, $Y_j^i$ is the ground-truth annotation, and $\ell(\cdot)$ includes cross-entropy for classification and smooth-$\mathcal{L}_1$ for localization.

\vspace{4pt}
\noindent{\bf{Routing Regularization:}}  
To prevent expert collapse and ensure balanced specialization, we introduce a routing regularization to penalizes deviation from uniform expert usage
\begin{equation}
    \mathcal{L}_{\text{route}} = \sum_{e=1}^{E_{G}} \left( \bar{p}_e - \frac{1}{E_{G}} \right)^2,
    \label{eq16}
\end{equation}
where $\bar{p}_e$ denotes the average selection probability of expert $e$ over the training set. We adopt this $L_2$ formulation instead of entropy maximization due to its simplicity and stable optimization behavior.
To further stabilize routing and expert specialization, we use a progressive training strategy that incrementally activates the components of HMR-Net.

\begin{itemize}
    \item We initially disable expert routing, region segmentation, and the CEM. All inputs are processed by a shared backbone and baseline detection head to learn general representations. Only $\mathcal{L}_{\text{det}}$ is used.
    
    \item After warm-up, we activate domain expert routing and cross-expert distillation. Inputs are routed to top-1 domain experts, and $\mathcal{L}_{\text{route}}$ and $\mathcal{L}_{\text{distill}}$ are added to have balanced and consistent specialization.

    \item In the final phase, scene region routing and the CEM are activated. Region segmentation is supervised by the coherence loss $\mathcal{L}_{\text{coh}}$, while visual-semantic alignment is optimized through $\mathcal{L}_{\text{align}}$.
\end{itemize}

\noindent{\bf{Total Loss:}}  
The full training is a weighted sum of the detection loss and additional terms:
\begin{equation}
    \mathcal{L}_{\text{total}} = \mathcal{L}_{\text{det}} + \lambda_1 \mathcal{L}_{\text{route}} + \lambda_2 \mathcal{L}_{\text{distill}}
    + \lambda_3 \mathcal{L}_{\text{align}} + \lambda_4 \mathcal{L}_{\text{coh}},
    \label{eq:total_loss}
\end{equation}
where $\lambda_1$, $\lambda_2$, $\lambda_3$, and $\lambda_4$ are weighting hyperparameters. During backpropagation, each expert receives gradients only from its assigned samples, enabling independent specialization (Algorithm~\ref{alg:hmr}). This staged training improves convergence stability by reducing early overfitting to suboptimal routing decisions and encouraging hierarchical specialization.

\section{Experiments}\label{sec4}
This section presents a comprehensive evaluation of the proposed HMR-Net. We evaluate its performance across a diverse set of benchmarks, focusing on three core capabilities: generalization across domains, specialization within complex scenes, and recognition of novel object categories. We conduct comparisons against existing state-of-the-art object detection models to highlight the benefits of modular specialization and structured routing. The experimental setup, evaluation protocols, and implementation details are described below.


\subsection{Experimental Setup}

\noindent {\bf {Datasets.}} We evaluate our model using four aerial object detection benchmarks:

\noindent {{DIOR}} \cite{li2020object} consists of 23,463 images across 20 object categories, including vehicles, buildings, aircraft, ships, and storage tanks. It covers multiple scenes, such as urban, rural, and industrial, captured at varying scales and resolutions.

\noindent {{DOTA-v1.0}} \cite{xia2018dota} contains high-resolution imagery showing complex spatial configurations and detailed scene compositions. It has 2,806 images of 15 object classes.

\noindent {{xView}} \cite{lam2018xview} is a large-scale dataset designed for detecting small and rare objects in overhead satellite imagery. It contains approximately 1 million labeled instances across 60 object categories, featuring both common classes and less frequent classes, making it valuable for testing the detection of sparsely represented objects.


\noindent {{NWPU VHR-10}} \cite{cheng2016learning} provides 800 high-resolution images of 10 object categories.


\begin{figure*}[t]
\centering
\includegraphics[width=1.65\columnwidth]{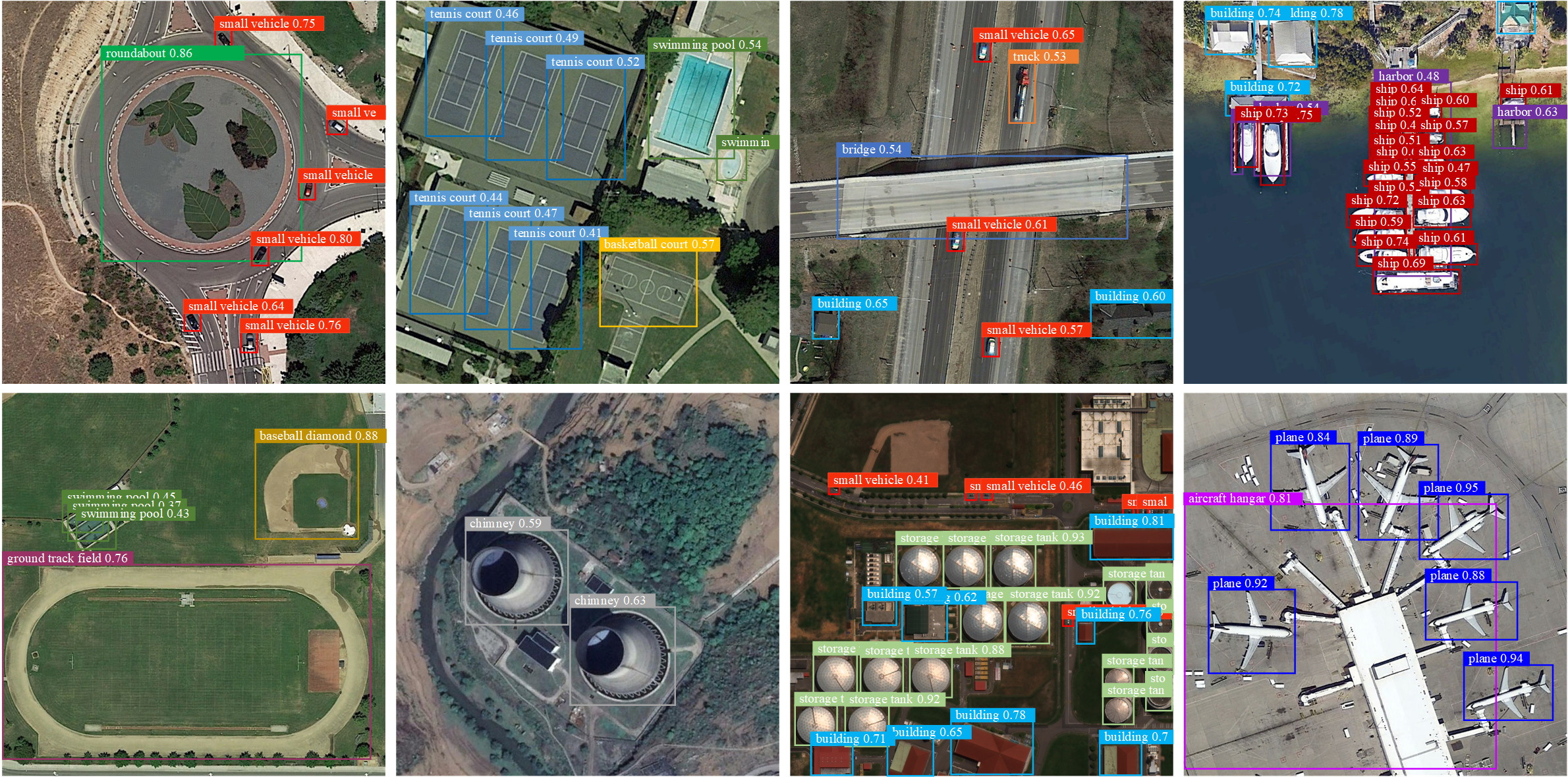} 
\vspace{-3pt}
\caption{\small{Representative detection results of HMR-Net on diverse aerial scenes and object categories.}}
\label{detection}
\end{figure*} 
\begin{table}[t]
\centering
\scriptsize
\setlength{\tabcolsep}{3.4pt}
\caption{{DIOR evaluation under multi-domain training.}}
\label{tab1}
\vspace{-3pt}

\begin{tabular}{
>{\raggedright\arraybackslash}p{1.7cm}
ccccccc
}
\toprule
Methods &
AP$\uparrow$ &
AP$_{50}\uparrow$ &
AP$_{75}\uparrow$ &
AP$_S\uparrow$ &
AP$_M\uparrow$ &
AP$_L\uparrow$ &
Avg \\
\midrule

FR-CNN \cite{ren2015faster}
& 27.53 & 51.26 & 28.11 & 21.43 & 28.65 & 47.38 & 34.05 \\

CR-CNN \cite{xu2022detecting}
& 28.90 & 52.78 & 28.65 & 22.17 & 29.40 & 49.04 & 35.16 \\

Me R-CNN \cite{lee2019me}
& 27.84 & 52.63 & 28.59 & 22.03 & 29.11 & 48.86 & 34.84 \\

RTMDet \cite{lyu2022rtmdet}
& 27.85 & 51.53 & 28.63 & 21.52 & 29.15 & 48.82 & 34.59 \\

H2RBox \cite{yang2022h2rbox}
& 31.42 & 55.19 & 31.27 & 26.44 & 35.82 & 51.69 & 38.64 \\

DAMEX \cite{jain2023damex}
& 30.46 & 55.84 & 31.02 & 22.73 & 35.64 & 52.14 & 37.97 \\

AFD \cite{shamsolmoali2023efficient}
& 30.82 & 54.22 & 30.26 & 24.72 & 32.51 & 50.16 & 37.12 \\

MoCaE \cite{oksuz2023mocae}
& 31.37 & 55.41 & 31.25 & 26.24 & 35.73 & 51.56 & 38.59 \\

ORFENet \cite{liu2024tiny}
& 33.19 & 56.45 & \underline{32.97}
& \textbf{27.53}
& 35.88
& 51.64
& 39.61 \\

TSC \cite{yu2026tsc}
& 33.12 & 56.50 & {32.84}
& \underline{27.46}
& 35.66
& 51.72
& 39.55 \\

LSKNet \cite{li2024lsknet}
& \underline{33.35}
& \textbf{56.88}
& 32.71
& {27.28}
& \underline{35.93}
& \underline{51.85}
& \underline{39.67} \\

\rowcolor{ForestGreen!15}
HMR-Net
& \textbf{33.48}
& \underline{56.61}
& \textbf{33.14}
& 27.20
& \textbf{35.96}
& \textbf{51.97}
& \textbf{39.72} \\

\bottomrule
\end{tabular}
\vspace{-4pt}
\end{table}
\noindent {\bf{Implementation Details.}}
For all experiments, we use HMR-Net with a Faster R-CNN R50-FPN head \cite{ren2015faster} pretrained on ImageNet. HMR-Net is detector-agnostic, as hierarchical routing produces a fused feature map $\mathbf{o}_j^i$ that can be used by either proposal-based or query-based detectors. Specifically, $\mathbf{o}_j^i$ is processed by an RPN and detection heads in Faster R-CNN, or used as the encoder input in DETR \cite{zhang2022dino}. Since routing is performed before the detection head, the same feature representation is shared across both settings.
We use six domain experts in all experiments. Rather than enforcing a one-to-one mapping between datasets and experts, the routing mechanism learns semantic groupings across datasets, enabling specialization based on visual content and improving generalization across shared object categories and scene types. All expert modules are implemented as lightweight networks consisting of three convolutional layers with ReLU activations and batch normalization. The choice of $E_G=6$ is supported by the ablation study in Fig.~\ref{ab-experts}, which shows the best balance between detection performance and expert utilization.

For the conditional expert, additional linear projection layers align visual embeddings with the CLIP semantic space. These layers are trained from scratch and operate on features produced by the hierarchical routing pipeline.

Training uses stochastic gradient descent with momentum 0.9, an initial learning rate of 0.001, and weight decay of $1\times10^{-4}$. Models are trained for 200 epochs, with the learning rate reduced by a factor of 0.1 at epochs 120 and 160. We set $\theta_R=1.0$. Training is performed on four RTX-3090 GPUs with a batch size of 8, while evaluation uses a single GPU. Standard data augmentations, including random flipping, scaling, and rotation, are applied during training.

All images are resized or tiled to a fixed resolution of $800\times800$. For DOTA and xView, we use non-overlapping tiling, whereas NWPU and DIOR are resized or padded to the target resolution. To ensure consistent evaluation, all annotations are converted to horizontal bounding boxes (HBB). At inference, Non-Maximum Suppression is applied with an IoU threshold of 0.5. Hyperparameters associated with routing regularization and prompt alignment are kept fixed across all experiments.
\noindent {\bf{Evaluation Protocols.}} We design comprehensive evaluation protocols to systematically validate our framework.

\begin{itemize}
\item {\it{Per-Dataset Evaluation:}} We evaluate the model on each dataset independently to assess its ability to adapt to domain-specific visual distributions and annotation schemas. This setting also evaluates the effectiveness of dataset-aware routing for expert specialization across diverse domains.
\begin{table}[t]
\centering
\scriptsize
\setlength{\tabcolsep}{3.4pt}
\caption{{DOTA evaluation under multi-domain training.}}
\label{tab2}
\vspace{-3pt}

\begin{tabular}{
>{\raggedright\arraybackslash}p{1.7cm}
ccccccc
}
\toprule
Methods &
AP$\uparrow$ &
AP$_{50}\uparrow$ &
AP$_{75}\uparrow$ &
AP$_S\uparrow$ &
AP$_M\uparrow$ &
AP$_L\uparrow$ &
Avg \\
\midrule

FR-CNN \cite{ren2015faster}
& 46.61 & 74.58 & 47.33 & 35.39 & 53.72 & 71.16 & 54.80 \\

CR-CNN \cite{xu2022detecting}
& 47.49 & 75.26 & 48.81 & 37.24 & 55.64 & 72.43 & 56.12 \\

Me R-CNN \cite{lee2019me}
& 48.64 & 77.48 & 51.13 & 38.26 & 57.85 & 72.59 & 57.66 \\

RTMDet \cite{lyu2022rtmdet}
& 51.43 & 80.42 & 56.35 & 42.46 & 60.28 & 73.79 & 60.78 \\

H2RBox \cite{yang2022h2rbox}
& 47.54 & 75.31 & 51.42 & 39.84 & 56.61 & 72.68 & 57.24 \\

DAMEX \cite{jain2023damex}
& 48.12 & 78.05 & 52.46 & 40.57 & 58.26 & 73.10 & 58.46 \\

AFD \cite{shamsolmoali2023efficient}
& 50.73 & 78.48 & 55.32 & 45.51 & 59.72 & 73.38 & 60.53 \\

MoCaE \cite{oksuz2023mocae}
& 50.56 & 81.03 & 55.64 & 43.38 & 61.70 & 74.89 & 61.20 \\

ORFENet \cite{liu2024tiny}
& 52.85
& 80.97
& \underline{58.84}
& \textbf{50.03}
& \underline{62.81}
& 75.37
& 63.49 \\

TSC \cite{yu2026tsc}
& 52.64
& 80.91
& {58.66}
& {49.62}
& {62.68}
& 75.27
& 63.30 \\

LSKNet \cite{li2024lsknet}
& \underline{53.17}
& \textbf{81.30}
& 58.71
& 49.59
& 62.74
& \underline{75.53}
& \underline{63.52} \\

\rowcolor{ForestGreen!15}
HMR-Net
& \textbf{53.23}
& \underline{81.24}
& \textbf{59.08}
& \underline{49.82}
& \textbf{63.18}
& \textbf{75.91}
& \textbf{63.74} \\

\bottomrule
\end{tabular}
\vspace{-5pt}
\end{table}

\vspace{2pt}
\item {\it{Cross-Domain Generalization:}} To evaluate generalization to unseen data distributions, we adopt a leave-one-dataset-out protocol, training on three datasets and testing on the remaining one. This setting reflects real-world scenarios where models encounter unseen geographic regions or imaging conditions and assesses the effectiveness of hierarchical routing for cross-domain transfer.

\vspace{2pt}
\item {\it{Open-Category Detection:}} We evaluate open-category detection by assessing the ability to recognize object classes excluded during training using textual prompts at inference time. This evaluates visual-semantic alignment and zero-shot detection performance without retraining. Following \cite{huang2022robust}, we adopt the predefined DIOR split. For DOTA, we use 11 seen and 4 unseen classes selected through semantic clustering. Textual prompts include both class names (e.g., airplane, harbour) and descriptive phrases (e.g., solar panel beside car). Up to 16 prompts are randomly sampled per image.
\end{itemize}

To evaluate performance across object scales, we report AP metrics stratified by object size. Following \cite{wang2021tiny, liu2024tiny}, object categories are defined using bounding box dimensions after image resizing: AP$_S$ for objects smaller than $32\times32$ pixels, AP$_M$ for objects between $32\times32$ and $64\times64$ pixels, and AP$_L$ for objects larger than $64\times64$ pixels. For open-category detection, following the standard zero-shot detection (ZSD) protocol \cite{bansal2018zero, huang2022robust}, we report Recall@100 (RE@100) at IoU thresholds of 0.4, 0.5, and 0.6, as well as mAP on test images containing only unseen categories.

\subsection{Comparison with State-of-the-Art}
We compare the performance of our proposed method against Faster R-CNN (FR-CNN) \cite{ren2015faster}, Cascade R-CNN (CR-CNN) \cite{xu2022detecting}, Me R-CNN \cite{lee2019me}, MoCaE \cite{oksuz2023mocae}, H2RBox \cite{yang2022h2rbox}, DAMEX \cite{jain2023damex}, RTMDet \cite{lyu2022rtmdet}, AFD \cite{shamsolmoali2023efficient}, ORFENet \cite{liu2024tiny}, TSC \cite{yu2026tsc} and LSKNet \cite{li2024lsknet}. For open-category (prompt-based) detection, we additionally compare against Detic \cite{zhou2022detecting}, ViLD \cite{gu2021open}, DetPro \cite{du2022learning}, AlignDet \cite{zareapoor2024learning} and CastDet \cite{li2026exploiting}. To ensure a fair comparison, we reproduce their official experiments by retraining all models using the same backbone architecture, training settings, and hardware environment as our method.

\subsubsection{Per-Dataset Evaluation} 
Models are trained on the combined training sets of all datasets and evaluated separately on each test set without fine-tuning. Fig. \ref{detection} shows representative detection results across datasets.

\noindent {\bf {(a) Evaluation on the DIOR.}}
As shown in Table \ref{tab1}, our model achieves the highest average performance ($39.72\%$), surpassing all competing methods. It also obtains the best AP$_{75}$ and strong performance across object scales, particularly for medium and large objects. These results highlight the effectiveness of HMR-Net in capturing both domain-level and spatial variations.

\begin{table}[t]
\centering
\scriptsize
\setlength{\tabcolsep}{3.4pt}
\caption{{xView evaluation under multi-domain training.}}
\label{tab3}
\vspace{-3pt}

\begin{tabular}{
>{\raggedright\arraybackslash}p{1.7cm}
ccccccc
}
\toprule
Methods &
AP$\uparrow$ &
AP$_{50}\uparrow$ &
AP$_{75}\uparrow$ &
AP$_S\uparrow$ &
AP$_M\uparrow$ &
AP$_L\uparrow$ &
Avg \\
\midrule

FR-CNN \cite{ren2015faster}
& 21.38 & 44.46 & 20.39 & 10.34 & 23.86 & 43.72 & 27.36 \\

CR-CNN \cite{xu2022detecting}
& 22.52 & 45.78 & 21.24 & 10.86 & 25.12 & 43.94 & 28.24 \\

Me R-CNN \cite{lee2019me}
& 23.17 & 45.91 & 21.83 & 12.37 & 25.06 & 44.10 & 28.74 \\

RTMDet \cite{lyu2022rtmdet}
& 23.43 & 46.03 & 22.16 & 12.78 & 25.14 & 44.23 & 28.96 \\

H2RBox \cite{yang2022h2rbox}
& 24.35 & 47.64 & 23.52 & 14.24 & 27.03 & 44.52 & 30.22 \\

DAMEX \cite{jain2023damex}
& 23.14 & 46.58 & 23.21 & 14.36 & 26.90 & 44.17 & 29.73 \\

AFD \cite{shamsolmoali2023efficient}
& 24.06 & 48.13 & 24.34 & 16.74 & 27.42 & 45.10 & 30.96 \\

MoCaE \cite{oksuz2023mocae}
& 24.82 & 48.76 & 25.17 & 17.85 & 27.89 & 45.51 & 31.67 \\

ORFENet \cite{liu2024tiny}
& \textbf{26.71}
& \underline{49.40}
& \underline{27.02}
& \textbf{19.84}
& \underline{28.75}
& 46.63
& \underline{33.06} \\

TSC \cite{yu2026tsc}
& {25.34}
& {48.85}
& {26.23}
& {17.52}
& {27.61}
& 45.94
& {31.92} \\

LSKNet \cite{li2024lsknet}
& 25.83
& 49.32
& 26.87
& 18.42
& 27.58
& \underline{46.86}
& 32.48 \\

\rowcolor{ForestGreen!15}
HMR-Net
& \underline{26.58}
& \textbf{49.61}
& \textbf{27.19}
& \underline{19.66}
& \textbf{28.79}
& \textbf{46.93}
& \textbf{33.12} \\

\bottomrule
\end{tabular}
\vspace{-5pt}
\end{table}

\begin{table}[b]
\centering
\scriptsize
\setlength{\tabcolsep}{3.4pt}
\caption{{NWPU evaluation under multi-domain training.}}
\label{tab4}
\vspace{-3pt}

\begin{tabular}{
>{\raggedright\arraybackslash}p{1.7cm}
ccccccc
}
\toprule
Methods &
AP$\uparrow$ &
AP$_{50}\uparrow$ &
AP$_{75}\uparrow$ &
AP$_S\uparrow$ &
AP$_M\uparrow$ &
AP$_L\uparrow$ &
Avg \\
\midrule

FR-CNN \cite{ren2015faster}
& 51.42 & 77.53 & 52.16 & 37.63 & 58.02 & 75.24 & 58.67 \\

CR-CNN \cite{xu2022detecting}
& 52.68 & 78.49 & 54.13 & 38.44 & 59.33 & 76.18 & 59.88 \\

Me R-CNN \cite{lee2019me}
& 54.47 & 80.16 & 57.86 & 41.66 & 61.63 & 77.68 & 62.24 \\

RTMDet \cite{lyu2022rtmdet}
& 54.34 & 80.02 & 57.91 & 41.52 & 61.45 & 77.40 & 62.11 \\

H2RBox \cite{yang2022h2rbox}
& 54.16 & 79.33 & 56.84 & 42.31 & 60.92 & 77.62 & 61.86 \\

DAMEX \cite{jain2023damex}
& 55.89 & 81.24 & 58.32 & 42.74 & 62.34 & 78.06 & 63.10 \\

AFD \cite{shamsolmoali2023efficient}
& 56.23 & 82.15 & 60.20 & 45.16 & 64.03 & 78.92 & 64.45 \\

MoCaE \cite{oksuz2023mocae}
& 58.16 & 83.34 & 62.37 & 46.31 & 65.19 & 79.84 & 65.87 \\

ORFENet \cite{liu2024tiny}
& 59.02
& {84.32}
& 63.05
& \textbf{51.13}
& \underline{66.82}
& 80.34
& 67.46 \\

TSC \cite{yu2026tsc}
& 58.72
& {83.85}
& 62.79
& {49.36}
& {66.21}
& 79.87
& 66.80 \\

LSKNet \cite{li2024lsknet}
& \textbf{59.76}
& \textbf{84.56}
& \underline{63.17}
& 50.87
& 66.73
& \underline{80.91}
& \underline{67.65} \\

\rowcolor{ForestGreen!15}
HMR-Net
& \underline{59.48}
& \underline{84.35}
& \textbf{63.91}
& \underline{50.98}
& \textbf{67.34}
& \textbf{81.24}
& \textbf{67.88} \\
\bottomrule
\end{tabular}
\vspace{-5pt}
\end{table}

\noindent {\bf {(b) Evaluation on the DOTA.}}
As Table \ref{tab2} shows, Our method achieves the highest overall performance with an average AP of $63.74\%$. It ranks first in AP$_{75}$, and robustness to medium and large object scales. While AP$_S$ ($49.82\%$) is slightly behind ORFENet ($50.03\%$), our model maintains balanced performance across all metrics, reflecting strong generalization across spatial structures. 

\noindent {\bf {(c) Evaluation on the xView.}}
xView presents significant challenges due to dense scenes and numerous small objects. As shown in Table \ref{tab3}, our model achieves the best overall performance ($33.12\%$). Despite a slightly lower AP$_S$ than ORFENet, it delivers more balanced performance across object scales and localization metrics.
\vspace{2pt}


\noindent {\bf {(d) Evaluations on the NWPU.}}
As shown in Table \ref{tab4}, all methods benefit from the relatively simple structure and larger objects in NWPU. However, our model achieves the best overall performance ($67.88\%$) and the highest AP$_{75}$, AP$_M$ and AP$_L$. These results demonstrate that hierarchical routing remains effective not only in complex scenes but also in more structured environments.

\begin{figure}[t]
\centering
\includegraphics[width=0.99\columnwidth]{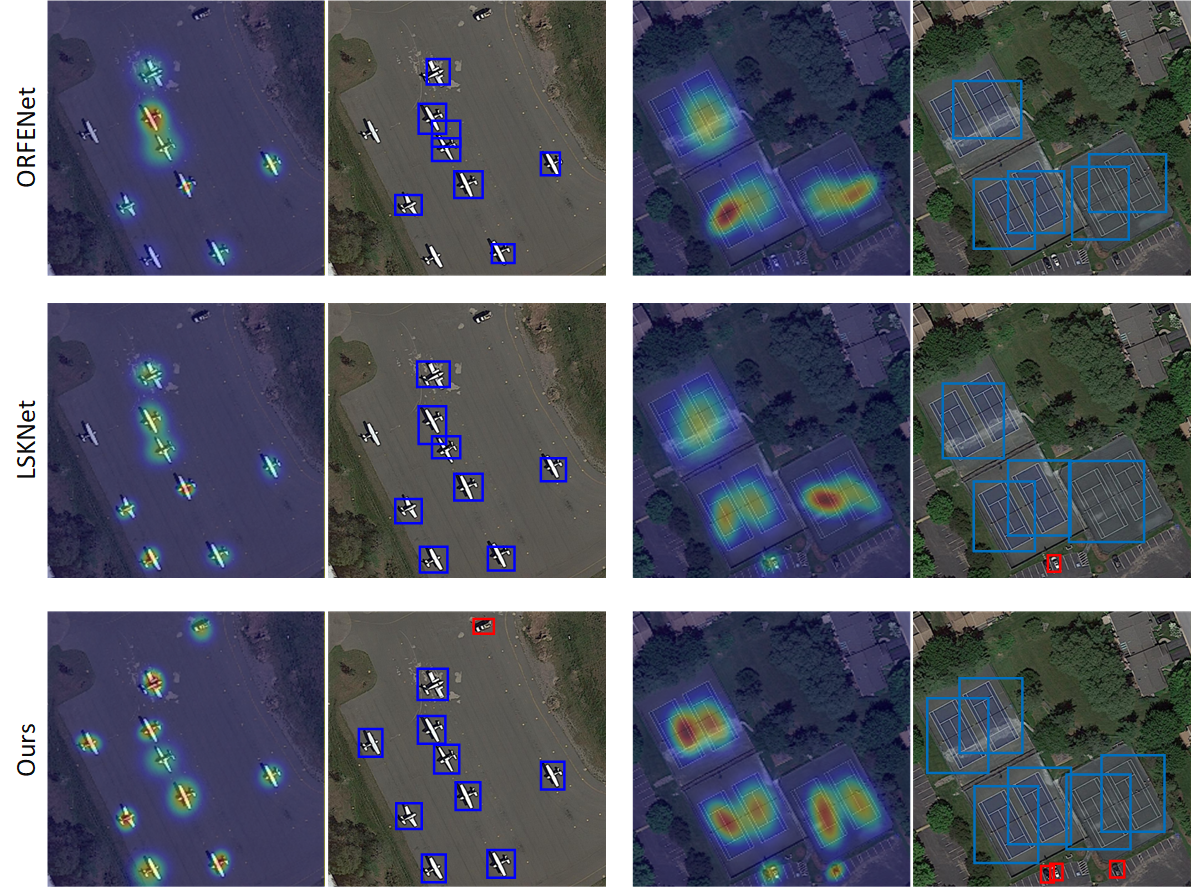} 
\vspace{-5pt}
\caption{\small{Detection results and feature map visualizations comparing HMR-Net against the two strongest baselines on DOTA. Our method shows more spatially focused representations and localizations.}}
\label{detection2}
\end{figure} 
\subsubsection{Cross-Domain Generalization} 
\mbox{}\\
\vspace{-11pt}

\noindent\textbf{(a) Generalization to DIOR.}
Models are trained on DOTA, xView, and NWPU, and evaluated on DIOR. As shown in Table \ref{tab5}, our model achieves the highest average precision ($35.21\%$), outperforming all baselines, including ORFENet and LSKNet. These results demonstrate the effectiveness of hierarchical routing in generalizing to unseen domains.
\begin{table}
\centering
\scriptsize
\setlength{\tabcolsep}{3.4pt}
\caption{{Cross-dataset generalization on the DIOR.}}
\label{tab5}
\vspace{-3pt}

\begin{tabular}{
>{\raggedright\arraybackslash}p{1.7cm}
ccccccc
}
\toprule
Methods &
AP$\uparrow$ &
AP$_{50}\uparrow$ &
AP$_{75}\uparrow$ &
AP$_S\uparrow$ &
AP$_M\uparrow$ &
AP$_L\uparrow$ &
Avg \\
\midrule

FR-CNN \cite{ren2015faster}
& 21.83 & 44.68 & 20.46 & 15.02 & 23.22 & 40.91 & 27.69 \\

CR-CNN \cite{xu2022detecting}
& 23.16 & 45.91 & 21.14 & 16.21 & 24.10 & 42.45 & 28.83 \\

Me R-CNN \cite{lee2019me}
& 24.53 & 46.68 & 21.98 & 17.35 & 26.12 & 43.67 & 30.06 \\

RTMDet \cite{lyu2022rtmdet}
& 22.75 & 44.84 & 20.83 & 15.16 & 23.96 & 41.89 & 28.24 \\

H2RBox \cite{yang2022h2rbox}
& 25.44 & 47.31 & 23.57 & 19.04 & 27.36 & 44.78 & 31.25 \\

DAMEX \cite{jain2023damex}
& 25.37 & 48.26 & 23.18 & 17.31 & 28.57 & 44.41 & 31.35 \\

AFD \cite{shamsolmoali2023efficient}
& 25.02 & 46.26 & 22.73 & 18.43 & 26.83 & 43.65 & 30.49 \\

MoCaE \cite{oksuz2023mocae}
& 26.14 & 47.85 & 24.67 & 19.80 & 28.41 & 44.83 & 31.95 \\

ORFENet \cite{liu2024tiny}
& 26.37
& 49.12
& \underline{25.38}
& \underline{21.01}
& 28.96
& 44.71
& 32.59 \\

TSC \cite{yu2026tsc}
& 26.14
& 48.87
& {24.69}
& {20.33}
& 28.36
& 44.12
& 32.08 \\

LSKNet \cite{li2024lsknet}
& \underline{26.59}
& \underline{49.34}
& 24.93
& 20.72
& \underline{29.13}
& \underline{45.10}
& \underline{32.63} \\

\rowcolor{ForestGreen!15}
HMR-Net
& \textbf{28.72}
& \textbf{52.20}
& \textbf{27.63}
& \textbf{22.79}
& \textbf{31.54}
& \textbf{48.36}
& \textbf{35.21} \\

\bottomrule
\end{tabular}
\vspace{-5pt}
\end{table}
\begin{table}[t]
\centering
\scriptsize
\setlength{\tabcolsep}{3.4pt}
\caption{{Cross-dataset generalization on the DOTA.}}
\label{tab6}
\vspace{-3pt}

\begin{tabular}{
>{\raggedright\arraybackslash}p{1.7cm}
ccccccc
}
\toprule
Methods &
AP$\uparrow$ &
AP$_{50}\uparrow$ &
AP$_{75}\uparrow$ &
AP$_S\uparrow$ &
AP$_M\uparrow$ &
AP$_L\uparrow$ &
Avg \\
\midrule

FR-CNN \cite{ren2015faster}
& 40.72 & 68.35 & 41.09 & 29.84 & 47.08 & 65.04 & 48.69 \\

CR-CNN \cite{xu2022detecting}
& 42.05 & 69.14 & 42.57 & 31.33 & 48.85 & 66.27 & 50.04 \\

Me R-CNN \cite{lee2019me}
& 44.32 & 72.79 & 47.51 & 34.85 & 50.62 & 67.08 & 52.86 \\

RTMDet \cite{lyu2022rtmdet}
& 44.28 & 72.60 & 47.68 & 35.17 & 50.76 & 67.01 & 52.92 \\

H2RBox \cite{yang2022h2rbox}
& 43.07 & 69.84 & 45.34 & 33.08 & 49.71 & 66.98 & 51.34 \\

DAMEX \cite{jain2023damex}
& 45.40 & 73.33 & 46.86 & 34.19 & 51.32 & 68.06 & 53.19 \\

AFD \cite{shamsolmoali2023efficient}
& 46.72 & 73.01 & 49.18 & 38.22 & 53.19 & 67.90 & 54.70 \\

MoCaE \cite{oksuz2023mocae}
& 47.58 & 74.53 & 50.91 & 40.85 & 54.67 & 68.42 & 56.16 \\

ORFENet \cite{liu2024tiny}
& 48.52
& 75.24
& \underline{52.58}
& \underline{43.11}
& \underline{56.32}
& 69.31
& 57.51 \\

TSC \cite{yu2026tsc}
& 47.43
& 74.36
& {50.81}
& {41.39}
& {55.27}
& 68.42
& 56.28 \\

LSKNet \cite{li2024lsknet}
& \underline{48.93}
& \underline{75.78}
& 52.41
& 42.47
& 56.25
& \underline{69.48}
& \underline{57.55} \\

\rowcolor{ForestGreen!15}
HMR-Net
& \textbf{50.38}
& \textbf{77.61}
& \textbf{54.05}
& \textbf{43.26}
& \textbf{58.23}
& \textbf{71.74}
& \textbf{59.21} \\

\bottomrule
\end{tabular}
\vspace{-5pt}
\end{table}

\noindent{\bf{(b) Generalization to DOTA.}}
Models are trained on DIOR, xView, and NWPU, and evaluated on DOTA. As shown in Table~\ref{tab6}, our model achieves the highest overall performance (Avg. 59.21\%) despite the challenges of precise localization and small object detection. These results demonstrate the effectiveness of HMR-Net in learning transferable representations for cross-domain generalization. As illustrated in Fig.~\ref{detection2}, our method produces more spatially aligned feature maps and more accurate detections than LSKNet.

\begin{table}[b]
\centering
\scriptsize
\setlength{\tabcolsep}{3.4pt}
\caption{{Cross-dataset generalization on the xView.}}
\label{tab7}
\vspace{-3pt}

\begin{tabular}{
>{\raggedright\arraybackslash}p{1.7cm}
ccccccc
}
\toprule
Methods &
AP$\uparrow$ &
AP$_{50}\uparrow$ &
AP$_{75}\uparrow$ &
AP$_S\uparrow$ &
AP$_M\uparrow$ &
AP$_L\uparrow$ &
Avg \\
\midrule

FR-CNN \cite{ren2015faster}
& 13.24 & 33.15 & 12.17 & 5.83 & 16.24 & 34.21 & 19.14 \\

CR-CNN \cite{xu2022detecting}
& 14.68 & 34.52 & 13.89 & 6.12 & 17.51 & 35.03 & 20.29 \\

Me R-CNN \cite{lee2019me}
& 14.61 & 34.46 & 13.71 & 6.07 & 17.44 & 35.10 & 20.23 \\

RTMDet \cite{lyu2022rtmdet}
& 15.75 & 36.31 & 14.75 & 7.24 & 18.19 & 35.78 & 21.34 \\

H2RBox \cite{yang2022h2rbox}
& 17.03 & 37.96 & 16.83 & 8.41 & 19.60 & 36.54 & 22.73 \\

DAMEX \cite{jain2023damex}
& 16.14 & 36.55 & 16.12 & 8.03 & 19.27 & 36.17 & 22.05 \\

AFD \cite{shamsolmoali2023efficient}
& 18.27 & 39.02 & 18.09 & 10.37 & 20.54 & 36.90 & 23.87 \\

MoCaE \cite{oksuz2023mocae}
& 19.64 & 41.23 & 19.93 & 11.69 & 21.65 & \underline{38.14} & 25.38 \\

ORFENet \cite{liu2024tiny}
& 20.53
& \underline{41.62}
& \underline{20.31}
& \textbf{12.52}
& \underline{22.49}
& 37.88
& \underline{25.90} \\

TSC \cite{yu2026tsc}
& 19.32
& {40.84}
& {19.83}
& {11.46}
& {21.68}
& 36.54
& {24.95} \\

LSKNet \cite{li2024lsknet}
& \underline{20.61}
& 40.96
& 20.05
& 11.87
& 21.84
& 38.05
& 25.56 \\

\rowcolor{ForestGreen!15}
HMR-Net
& \textbf{21.40}
& \textbf{42.36}
& \textbf{20.74}
& \underline{12.38}
& \textbf{23.14}
& \textbf{39.51}
& \textbf{26.59} \\

\bottomrule
\end{tabular}
\vspace{-5pt}
\end{table}
\vspace{2pt}
\noindent{\bf{(c) Generalization to xView.}}
Models are trained on DIOR, DOTA, and NWPU, and evaluated on xView to assess robustness under severe scale imbalance and dense small-object scenes. As shown in Table~\ref{tab7}, all methods experience notable performance degradation. Nevertheless, our model achieves the best overall performance, demonstrating the effectiveness of hierarchical routing in handling large domain shifts and complex scene structures.

\begin{table}[t]
\centering
\scriptsize
\setlength{\tabcolsep}{3.4pt}
\caption{{Cross-dataset generalization on NWPU.}}
\label{tab8}
\vspace{-3pt}

\begin{tabular}{
>{\raggedright\arraybackslash}p{1.7cm}
ccccccc
}
\toprule
Methods &
AP$\uparrow$ &
AP$_{50}\uparrow$ &
AP$_{75}\uparrow$ &
AP$_S\uparrow$ &
AP$_M\uparrow$ &
AP$_L\uparrow$ &
Avg \\
\midrule

FR-CNN \cite{ren2015faster}
& 48.24 & 74.92 & 49.08 & 35.26 & 54.77 & 72.25 & 55.76 \\

CR-CNN \cite{xu2022detecting}
& 49.37 & 75.88 & 50.46 & 36.41 & 56.10 & 73.42 & 56.94 \\

Me R-CNN \cite{lee2019me}
& 49.28 & 75.93 & 50.34 & 36.29 & 55.97 & 73.40 & 56.87 \\

RTMDet \cite{lyu2022rtmdet}
& 51.02 & 77.35 & 54.92 & 39.60 & 58.34 & 74.08 & 59.22 \\

H2RBox \cite{yang2022h2rbox}
& 50.89 & 76.70 & 53.64 & 41.22 & 58.92 & 74.59 & 59.33 \\

DAMEX \cite{jain2023damex}
& 52.16 & 78.26 & 55.12 & 40.97 & 57.20 & 74.11 & 59.64 \\

AFD \cite{shamsolmoali2023efficient}
& 53.08 & 79.60 & 57.06 & 43.24 & 60.84 & 76.08 & 61.65 \\

MoCaE \cite{oksuz2023mocae}
& 55.21 & 81.19 & 60.28 & 47.12 & 62.95 & 77.40 & 64.03 \\

ORFENet \cite{liu2024tiny}
& \underline{56.10}
& 81.45
& \underline{60.71}
& \underline{48.36}
& \underline{63.56}
& 77.31
& \underline{64.58} \\

TSC \cite{yu2026tsc}
& {55.32}
& 81.04
& {60.09}
& {47.41}
& {62.91}
& 77.05
& {63.97} \\

LSKNet \cite{li2024lsknet}
& 55.74
& \underline{81.61}
& 60.45
& 47.93
& 63.42
& \underline{78.02}
& 64.53 \\

\rowcolor{ForestGreen!15}
HMR-Net
& \textbf{56.86}
& \textbf{82.54}
& \textbf{61.24}
& \textbf{48.57}
& \textbf{64.43}
& \textbf{78.71}
& \textbf{65.39} \\

\bottomrule
\end{tabular}
\vspace{-5pt}
\end{table}

\vspace{2pt}
\noindent{\bf{(d) Generalization to NWPU.}}
Models are trained on DIOR, DOTA, and xView, and evaluated on NWPU. As shown in Table~\ref{tab8}, our model outperforms all baselines, demonstrating strong generalization and localization performance in structured scenes with larger objects and less clutter.

Across all cross-domain evaluations, our model consistently achieves the best performance, demonstrating robust generalization to unseen domains with diverse characteristics. These results highlight the effectiveness of HMR-Net as a transferable framework for aerial object detection.
\subsubsection{Open-Category Detection}
\begin{figure}
\centering
\includegraphics[width=0.99\columnwidth]{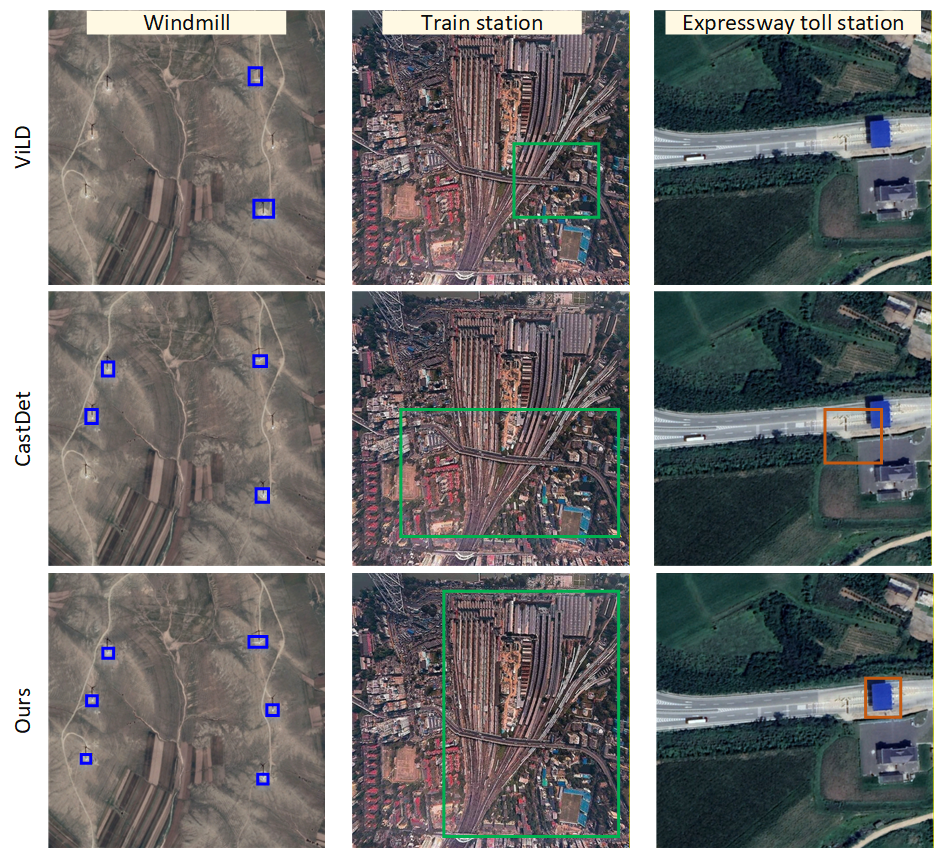} 
\vspace{-5pt}
\caption{\small{Qualitative results for unseen objects. Our method detects novel categories with improved precision and semantic alignment, compared to CastDet.}}
\label{ov-detection}
\end{figure}

Table \ref{tab9} compares our method with existing open-category detectors on DIOR and DOTA under the ZSD setting. Our model achieves the highest mAP on both datasets, reaching 16.42\% on DIOR and 9.78\% on DOTA, while also achieving strong RE@100 performance across IoU thresholds. These results demonstrate the effectiveness of the proposed visual-semantic alignment mechanism for detecting unseen categories in aerial imagery. Representative results are shown in Fig. \ref{ov-detection}.

\begin{table}
\centering
\scriptsize
\setlength{\tabcolsep}{3.2pt}
\caption{{Comparison with state-of-the-art methods under ZSD setting on the DIOR and DOTA. Performance is reported in terms of RE@100 at multiple IoU thresholds and mAP.}}
\label{tab9}
\vspace{-3pt}

\begin{tabular}{l|ccc|c|ccc|c}
\toprule

\multirow{3}{*}{\textbf{Method}}
& \multicolumn{4}{c|}{DIOR}
& \multicolumn{4}{c}{DOTA} \\

\cmidrule(lr){2-5}
\cmidrule(lr){6-9}

& \multicolumn{3}{c|}{RE@100}
& mAP
& \multicolumn{3}{c|}{RE@100}
& mAP \\

& IoU=.4 & IoU=.5 & IoU=.6 &
& IoU=.4 & IoU=.5 & IoU=.6 & \\

\midrule

ViLD \cite{gu2021open}
& 34.68 & 30.83 & 28.34 & 11.55
& 32.60 & 26.52 & 21.73 & 5.29 \\

Detic \cite{zhou2022detecting}
& 36.06 & 33.29 & 30.53 & 13.31
& 35.14 & 30.89 & 26.34 & 6.57 \\

DetPro \cite{du2022learning}
& 36.42 & 34.15 & \underline{31.72} & 15.45
& \underline{37.23} & 33.69 & 29.14 & 8.91 \\

AlignDet \cite{zareapoor2024learning}
& 37.75 & 34.81 & 31.19 & 15.76
& 36.34 & 34.28 & 29.57 & 9.12 \\

CastDet \cite{li2026exploiting}
& \underline{37.93}
& \textbf{35.03}
& 31.24
& \underline{16.11}
& 36.52
& \textbf{34.86}
& \underline{29.74}
& 9.35 \\

\rowcolor{ForestGreen!15}
Ours
& \textbf{38.21}
& \underline{34.84}
& \textbf{31.77}
& \textbf{16.42}
& \textbf{37.56}
& \underline{34.63}
& \textbf{29.95}
& \textbf{9.78} \\

\bottomrule
\end{tabular}
\vspace{-5pt}
\end{table}

\begin{figure}
\centering
\includegraphics[width=0.98\columnwidth]{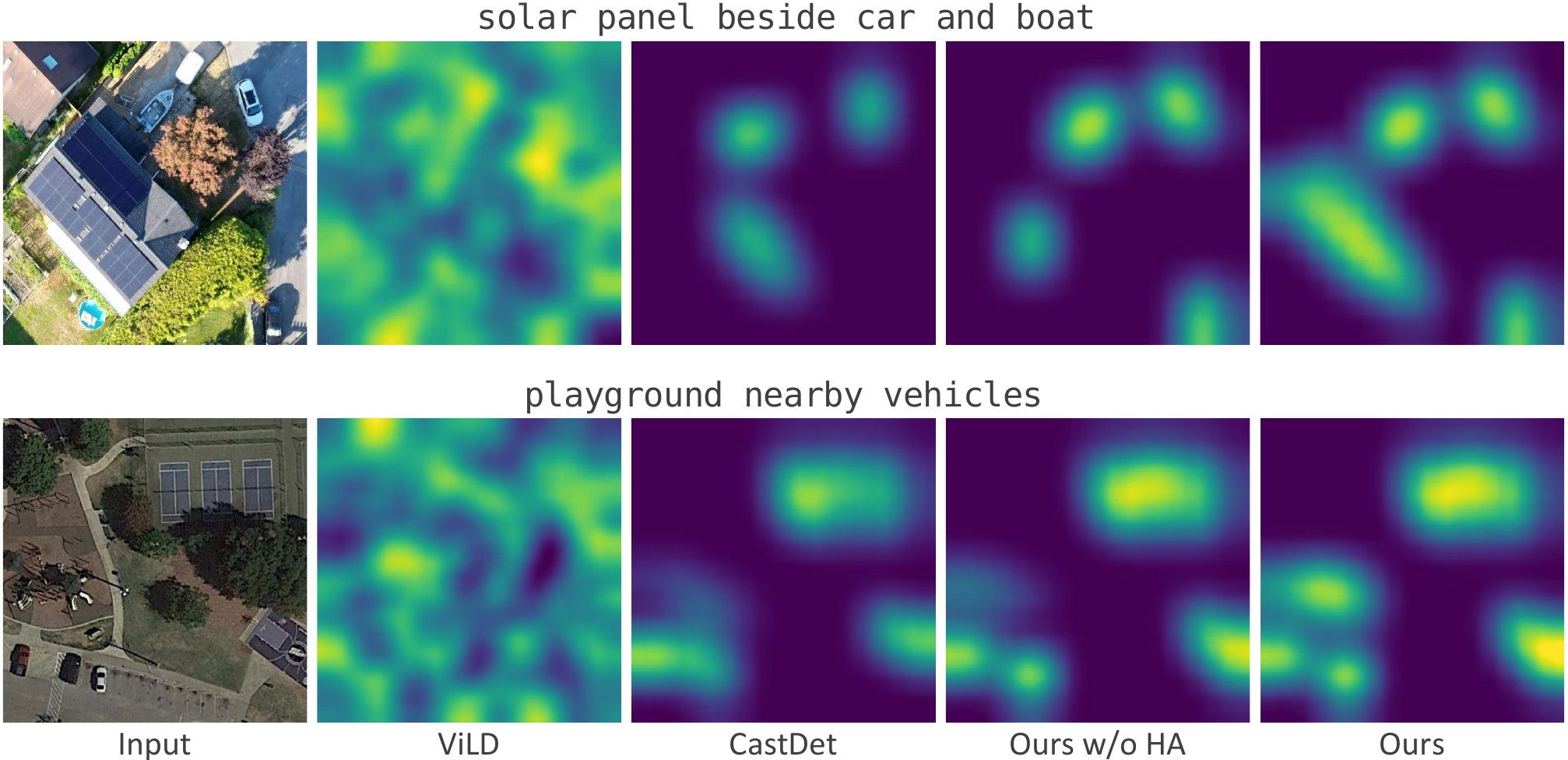} 
\vspace{-6pt}
\caption{\small{Visualization results for Zero-shot inference. In both examples, ViLD fails to distinguish the referent structure,
while CastDet and our model with bipartite matching localize partial entities. Our model with Hungarian algorithm (HA) shows finegrained attention and accurately interprets the entire prompt. Objects like “solar panel” or “playground” are not labeled in training,
but are localized through our textual alignment process.}}
\label{at-map}
\end{figure} 
\begin{table}
\centering
\scriptsize
\setlength{\tabcolsep}{3.1pt}
\caption{{Ablation study on DOTA and xView comparing the contributions of domain and scene expert routings.}}
\label{tab10}
\vspace{-3pt}

\begin{tabular}{l|cccc|cccc}
\toprule

\multirow{2}{*}{\textbf{Method}}
& \multicolumn{4}{c|}{DOTA}
& \multicolumn{4}{c}{xView} \\

\cmidrule(lr){2-5}
\cmidrule(lr){6-9}

& AP$_{50}$ & AP$_S$ & AP$_M$ & AP$_L$
& AP$_{50}$ & AP$_S$ & AP$_M$ & AP$_L$ \\

\midrule

HMR-Net (Domain only)
& 77.61 & 44.15 & 59.78 & 73.26
& 45.83 & 13.59 & 25.32 & 44.39 \\

HMR-Net (Scene only)
& 79.43 & 47.06 & 62.18 & 74.42
& 46.12 & 15.75 & 26.74 & 44.76 \\

\rowcolor{ForestGreen!15}
HMR-Net
& \textbf{81.24}
& \textbf{49.82}
& \textbf{63.18}
& \textbf{75.91}
& \textbf{49.57}
& \textbf{19.59}
& \textbf{28.75}
& \textbf{46.92} \\

\bottomrule
\end{tabular}
\vspace{-5pt}
\end{table}
\begin{figure}[t]
\centering
\includegraphics[width=0.98\columnwidth]{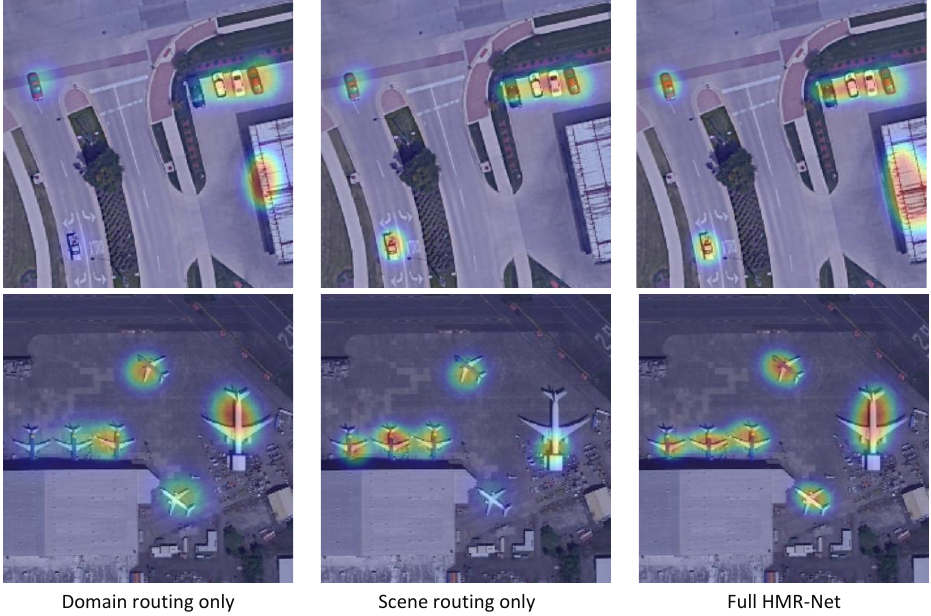} 
\vspace{-5pt}
\caption{\small{Comparison of feature maps generated by domain routing, scene routing, and their combined approach.}}
\label{attention}
\end{figure} 
Fig. \ref{at-map} shows zero-shot inference results, where each row shows a ground-truth image and the corresponding activation map generated from a descriptive prompt. ViLD produces diffuse activations across the image, reflecting its limited ability to handle unseen or compositional categories. CastDet generates more spatially localized responses but struggles with categories absent from the training annotations. Removing the Hungarian algorithm and using greedy top-1 matching further degrades localization by failing to capture fine-grained region-prompt correspondences. In contrast, HMR-Net uses optimal bipartite matching through the Hungarian algorithm, producing more accurate and concentrated activations for both seen and unseen categories, while better capturing the semantic relationships described by the prompts.
\subsection{Ablation Studies}
We conduct ablation studies to evaluate the contribution of each component of HMR-Net to both object detection and open-category recognition.

\subsubsection{Hierarchical Modular Routing}
To assess the contribution of each routing stage, we compare three configurations: domain-only, scene-only, and the full model. The domain-only variant performs expert selection using dataset-level embeddings, while the scene-only variant relies on content-aware routing for region-level specialization. The full model combines both stages through hierarchical routing.

As shown in Table~\ref{tab10} and Fig.~\ref{attention}, both routing components improve detection performance. Domain routing benefits from dataset-level priors but is less effective for small and medium-sized objects, reflecting its limited ability to capture fine-grained spatial variations. Scene routing improves performance across object scales, particularly on complex datasets such as xView, by enabling content-driven specialization. However, without domain routing, expert assignment becomes less reliable under strong domain shifts.
\begin{figure}[t]
\centering
\includegraphics[width=0.99\columnwidth]{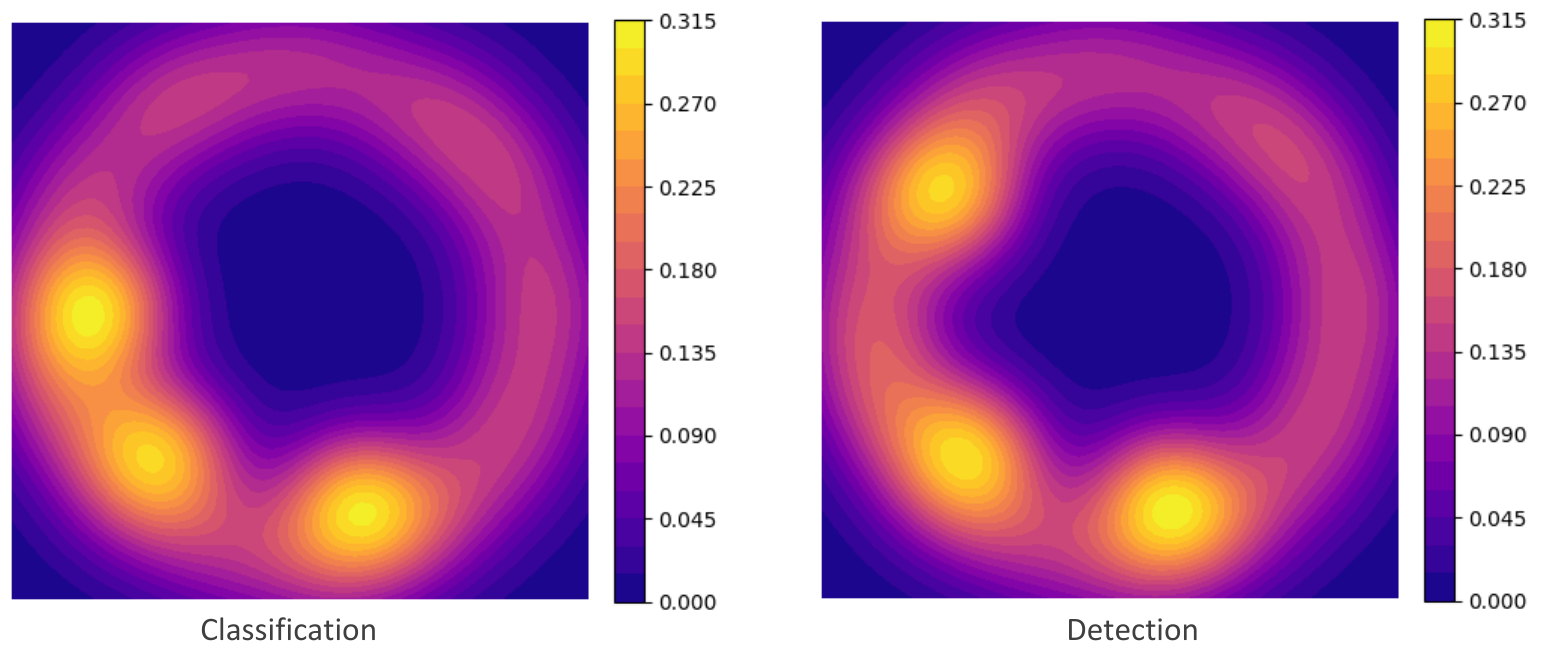} 
\vspace{-10pt}
\caption{\small{Evaluation of router representation collapse. Lighter regions indicate higher attention. Uniform heatmaps suggest balanced expert selection and reduced collapse.}}
\label{collapse}
\end{figure} 

The full HMR-Net consistently outperforms both variants across all datasets, demonstrating the benefit of combining domain-aware routing with fine-grained spatial specialization. In particular, it achieves 81.24 AP$_{50}$ on DOTA and 49.57 AP on xView, with consistent improvements across object scales. These results confirm the effectiveness of hierarchical routing in capturing various scene variations.

We further evaluate the effect of the proposed regularization strategy on routing stability and representation collapse. Fig.~\ref{collapse} visualizes router representations using Gaussian kernel density estimation. The resulting heatmaps show well-distributed hidden states for both classification and detection, indicating balanced expert utilization and a reduced risk of representation collapse.

\begin{figure}[t]
\centering
\includegraphics[width=0.99\columnwidth]{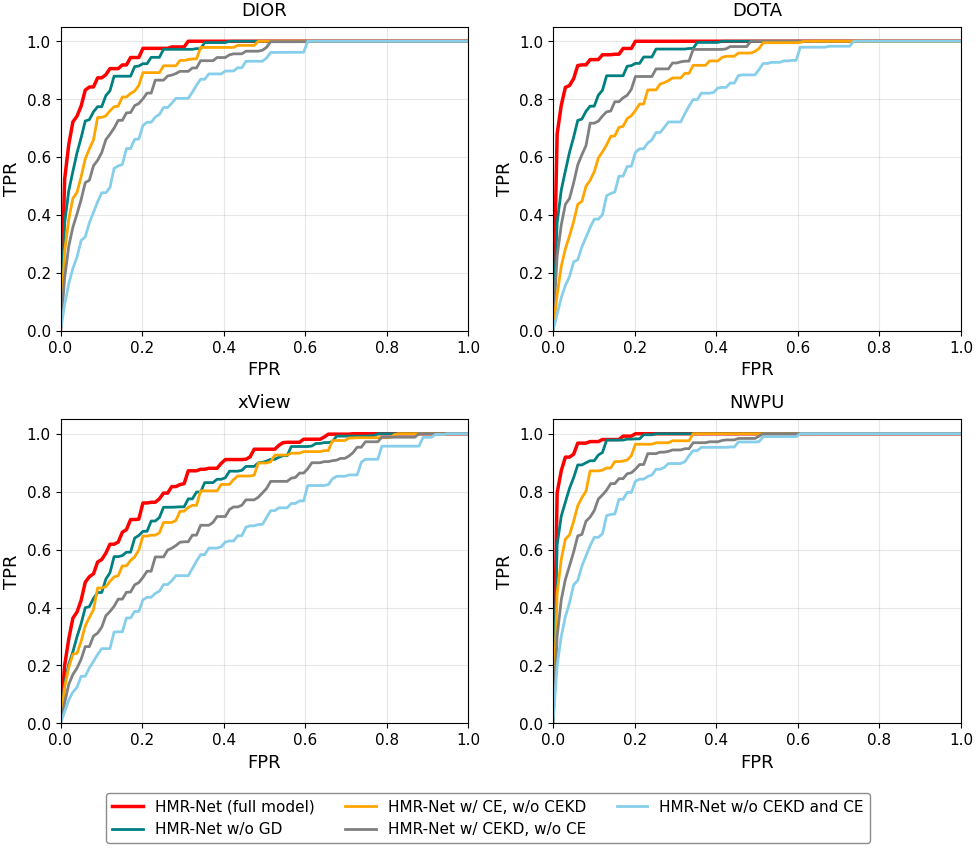} 
\vspace{-10pt}
\caption{\small{ROC curves on four evaluation datasets, illustrating the individual contributions of CEKD, Complexity Estimation (CE), and Global Distillation (GD). The full HMR-Net consistently outperforms all ablated variants across datasets, confirming the complementary and generalizable benefits of each component.}}
\label{tp-fp}
\end{figure} 
\begin{figure}
\centering
\includegraphics[width=0.99\columnwidth]{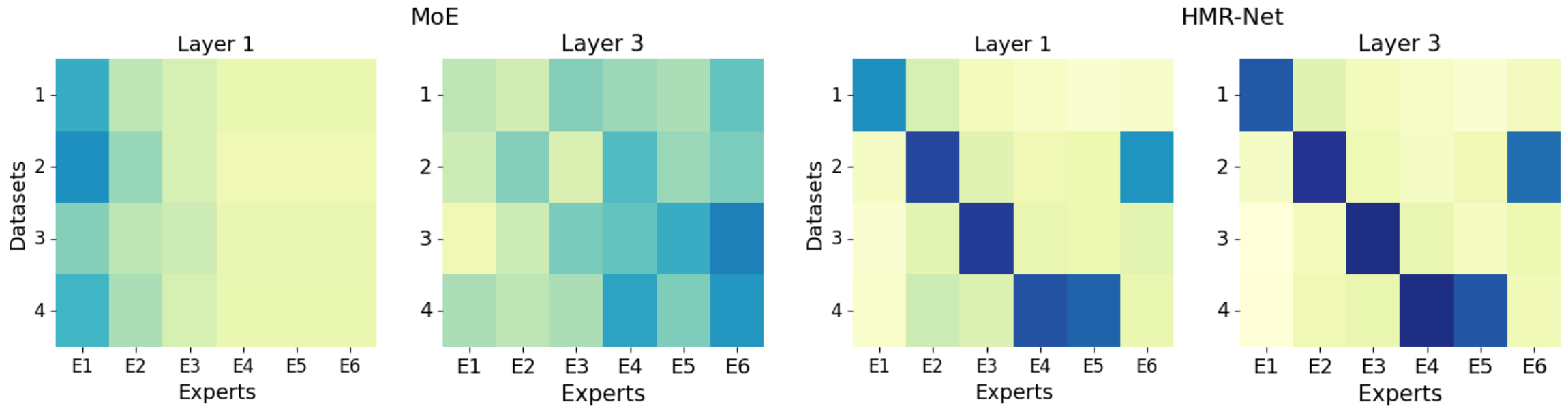} 
\vspace{-10pt}
\caption{\small{Expert usage patterns across layers 1 and 3 for vanilla MoE and HMR-Net. Unlike MoE, which distributes inputs diffusely across experts, HMR-Net consistently routes each dataset to dedicated experts, producing structured and interpretable specialization from early layers onward.}}
\label{expert}
\end{figure} 

To evaluate the contributions of CEKD, Complexity Estimation, and Global Distillation, we conduct an ablation study (see Fig.~\ref{tp-fp}). The full HMR-Net consistently achieves the best performance, demonstrating the complementary benefits of expert consistency, adaptive region specialization, and domain expert alignment.
Removing CE leads to moderate performance degradation, indicating the importance of region-aware specialization. Removing CEKD causes a larger drop, particularly at low false positive rates, highlighting the role of cross-expert regularization in stable expert selection. Ablating GD further reduces performance, confirming its contribution to aligning domain experts and supporting effective scene specialization. The model without all three components performs worst on both datasets, underscoring the importance of these modules for reliable expert learning.

\begin{table}
\centering
\scriptsize
\setlength{\tabcolsep}{2pt}
\caption{{Ablation on detector head: Faster R-CNN vs. DETR.}}
\label{detr-compat}
\vspace{-3pt}

\begin{tabular}{lcccccc}
\toprule

Methods &
AP$_{50}\uparrow$ &
AP$_{75}\uparrow$ &
AP$_S\uparrow$ &
AP$_M\uparrow$ &
AP$_L\uparrow$ &
Params (M) \\

\midrule

Ours w Faster R-CNN
& 52.20
& 27.63
& 22.79
& 31.54
& 48.36
& 59 \\

Ours w DINO-DETR
& 51.97
& 27.40
& 20.33
& 30.72
& 48.49
& 71 \\

\bottomrule
\end{tabular}
\vspace{-5pt}
\end{table}
\begin{figure}
\centering
\includegraphics[width=0.99 \columnwidth]{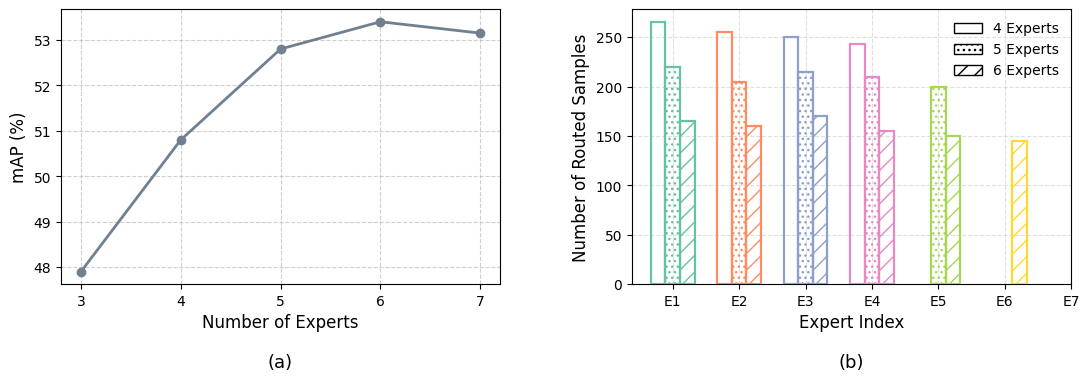} 
\vspace{-10pt}
\caption{\small{Impact of the number of domain experts $E_G$ when training on the 4-dataset mixture, reported on DOTA.
(a) mAP as the number of domain experts increases.
(b) Expert utilization across different expert configurations. The routing distribution becomes more balanced as the number of experts increases, with all experts actively contributing in the six-expert setting.}}
\label{ab-experts}
\end{figure} 

Fig.~\ref{expert} compares expert utilization in vanilla MoE \cite{puigcerver2023sparse} and HMR-Net across four datasets. The MoE heatmap exhibits a diffuse distribution, indicating that samples are distributed across experts without clear dataset-specific specialization. In contrast, HMR-Net produces a structured utilization pattern, with datasets consistently routed to specific experts. This specialization improves expert diversity and contributes to stronger cross-dataset generalization.

\noindent{\bf{Proposal-based vs. query-based detectors.}}
To verify that HMR-Net is detector-agnostic, we replace the proposal-based head with a DETR head while keeping the routing framework unchanged. The fused feature map $\mathbf{o}_j^i$ is used as the transformer encoder input, and detections are produced using the standard DETR matching loss \cite{zhang2022dino}. As shown in Table~\ref{detr-compat}, HMR-Net achieves strong performance with both Faster R-CNN and DETR, demonstrating its compatibility with both proposal-based and query-based detectors.

Additionally, Fig.~\ref{ab-experts} evaluates the effect of the number of domain experts on detection performance and expert utilization. As shown in Fig.~\ref{ab-experts}(a), increasing the number of experts from three to six consistently improves mAP, indicating enhanced specialization. Beyond six experts, the gains saturate and slightly decrease, suggesting diminishing returns from additional capacity.
Fig.~\ref{ab-experts}(b) shows that expert utilization becomes more balanced as the number of experts increases. With six experts, all modules are actively used, indicating effective capacity utilization. These results suggest that six domain experts provide the best trade-off between detection accuracy and model complexity.
\begin{figure}
\centering
\includegraphics[width=0.96 \columnwidth]{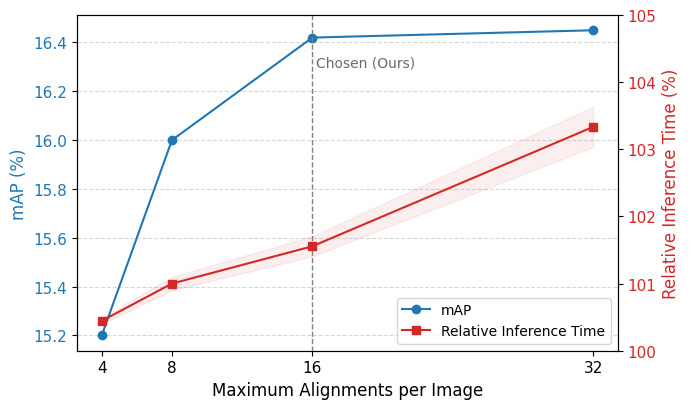} 
\vspace{-5pt}
\caption{\small{Performance vs computation for varying numbers of alignments. The blue line denotes mAP and the red line denotes relative inference time. The shaded band shows the min–max overhead across runs. Using 16 alignments per image achieves substantial accuracy gains with only a minor latency increase.}}
\label{N-image}
\end{figure} 

Fig.~\ref{N-image} illustrates the impact of the number of visual-text alignments in the CEM. Increasing the number of alignments improves detection accuracy up to 16 alignments, after which gains become marginal. Since the corresponding runtime increase remains small, 16 alignments provide an effective balance between accuracy and computational cost.

\section{Conclusion}\label{sec5}
In this paper, we introduced Hierarchical Modular Routing Network (HMR-Net), a modular framework that enables structured specialization across datasets and within complex aerial scenes. Through domain-level routing, scene-level routing, and a conditional expert module for open-category detection, HMR-Net achieves strong adaptability and generalization. Extensive experiments demonstrate consistent improvements over existing methods in multi-domain detection, cross-domain generalization, and zero-shot object detection. These results establish HMR-Net as an effective and scalable framework for remote sensing object detection. Future work will investigate adaptive expert pruning to improve efficiency while maintaining expert diversity and specialization.

\footnotesize{
\bibliographystyle{IEEEtran}
\bibliography{IEEE-TGRS}}

\end{document}